\documentstyle[mkstyle,twoside]{article}
\title{
Dempsterian-Shaferian Belief Network From Data}

\newcommand{\rDef}[1]{Def.\ref{#1}}

\newcommand{\V}{{\bf V }}
\newcommand{\SS}{{\bf S }}
\newcommand{\Alpha}{\mbox{\bf A }}
\newcommand{\Bem}[1]{}

\version{03}

\begin{document}

\machetitel

\setlength{\unitlength}{1cm}
\newcommand{\AbbEins}
{
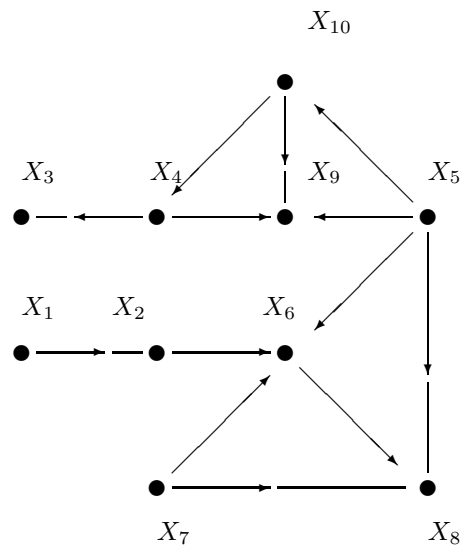
\begin{figure}
\begin{center}
\begin{picture}(8,8.3)
\put(3,2){\circle*{0.2}} 
\put(3,1.3){$X_7$}
\put(3.2,2){\vector(1,0){1.3}}
\put(4.6,2){\line(1,0){1.7}}
\put(3.2,2.2){\vector(1, 1){1.3}}    
\put(6.6,2){\circle*{0.2}}
\put(6.6,1.3){$X_8$}
\put(3,5.6){\circle*{0.2}}
\put(2.9,6.1){$X_4$} 
\put(2.8,5.6){\vector(-1,0){0.9}}  
\put(1.8,5.6){\line(-1,0){0.4}}    
\put(3.2,5.6){\vector(1,0){1.3}}
\put(4.7,7.4){\circle*{0.2}} 
\put(5,8.1){$X_{10}$} 
\put(4.5,7.2){\vector(-1,-1){1.3}}    
\put(4.7,7.2){\vector(0,-1){0.9}}
\put(4.7,6.2){\line(0,-1){0.4}}
\put(4.7,5.6){\circle*{0.2}} 
\put(5,6.1){$X_9$}
\put(6.6,5.6){\circle*{0.2}}
\put(6.6,6.1){$X_5$}
\put(6.4,5.4){\vector(-1, -1){1.3}}    
\put(6.6,5.4){\vector(0,-1){1.9}}
\put(6.6,3.4){\line(0,-1){1.2}}
\put(6.4,5.6){\vector(-1,0){1.3}}
\put(6.4,5.8){\vector(-1,1){1.3}}    
\put(1.2,5.6){\circle*{0.2}}
\put(1.2,6.1){$X_3$}
\put(1.2,3.8){\circle*{0.2}}
\put(1.2,4.3){$X_1$}
\put(1.4,3.8){\vector(1,0){0.9}}
\put(2.4,3.8){\line(1,0){0.4}}
\put(3,3.8){\circle*{0.2}} 
\put(2.4,4.3){$X_2$}
\put(3.2,3.8){\vector(1,0){1.3}}
\put(4.7,3.8){\circle*{0.2}} 
\put(4.4,4.3){$X_6$}
\put(4.9,3.6){\vector(1,-1){1.3}}    
\end{picture}
\end{center}
\caption{An Example of a Directed Acyclic Graph (dag)}  \label{abbeins}
\end{figure}
} 
%
\newcommand{\AbbZwei}
{
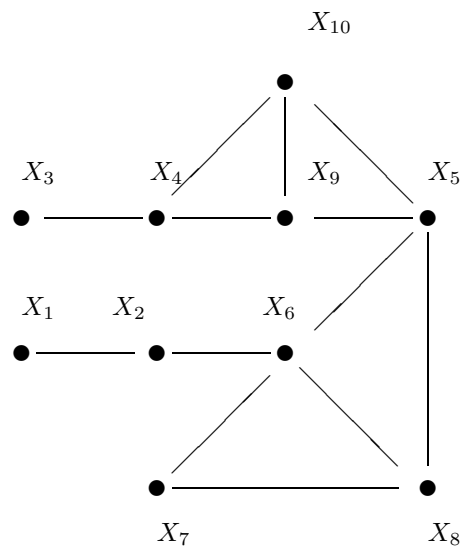
\begin{figure}
\begin{center}
\begin{picture}(8,8.3)
\put(3,2){\circle*{0.2}} 
\put(3,1.3){$X_7$}
\put(3.2,2){\line(1,0){3.0}}
\put(3.2,2.2){\line(1, 1){1.3}}    
\put(6.6,2){\circle*{0.2}}
\put(6.6,1.3){$X_8$}
\put(3,5.6){\circle*{0.2}}
\put(2.9,6.1){$X_4$} 
\put(2.8,5.6){\line(-1,0){1.3}}  
\put(3.2,5.6){\line(1,0){1.3}}
\put(4.7,7.4){\circle*{0.2}} 
\put(5,8.1){$X_{10}$} 
\put(4.5,7.2){\line(-1,-1){1.3}}    
\put(4.7,7.2){\line(0,-1){1.3}}
\put(4.7,5.6){\circle*{0.2}} 
\put(5,6.1){$X_9$}
\put(6.6,5.6){\circle*{0.2}}
\put(6.6,6.1){$X_5$}
\put(6.4,5.4){\line(-1, -1){1.3}}    
\put(6.6,5.4){\line(0,-1){3.1}}
\put(6.4,5.6){\line(-1,0){1.3}}
\put(6.4,5.8){\line(-1,1){1.3}}    
\put(1.2,5.6){\circle*{0.2}}
\put(1.2,6.1){$X_3$}
\put(1.2,3.8){\circle*{0.2}}
\put(1.2,4.3){$X_1$}
\put(1.4,3.8){\line(1,0){1.3}}
\put(3,3.8){\circle*{0.2}} 
\put(2.4,4.3){$X_2$}
\put(3.2,3.8){\line(1,0){1.3}}
\put(4.7,3.8){\circle*{0.2}} 
\put(4.4,4.3){$X_6$}
\put(4.9,3.6){\line(1,-1){1.3}}    
\end{picture}
\end{center}
\caption{An undirected graph obtained by application of Principle I}
\label{abbzwei}
\end{figure}
} 
%
%
\newcommand{\AbbDrei}
{
\begin{figure}
\begin{center}
\begin{picture}(8,8.3)
\put(3,2){\circle*{0.2}} 
\put(3,1.3){$X_7$}
\put(3.2,2){\line(1,0){3.0}}
\put(3.2,2.2){\vector(1, 1){1.3}}    
\put(6.6,2){\circle*{0.2}}
\put(6.6,1.3){$X_8$}
\put(3,5.6){\circle*{0.2}}
\put(2.8,5.6){\line(-1,0){1.3}}  
\put(2.9,6.1){$X_4$} 
\put(3.2,5.6){\vector(1,0){1.3}}   
\put(4.7,7.4){\circle*{0.2}} 
\put(5,8.1){$X_{10}$} 
\put(4.5,7.2){\line(-1,-1){1.3}}    
\put(4.7,7.2){\line(0,-1){1.3}}   
\put(4.7,5.6){\circle*{0.2}} 
\put(5,6.1){$X_9$}
\put(6.6,5.6){\circle*{0.2}}
\put(6.6,6.1){$X_5$}
\put(6.4,5.4){\vector(-1, -1){1.3}}    
\put(6.6,3.4){\line(0,-1){1.2}}    
\put(6.6,5.4){\line(0,-1){3.1}}   
\put(6.4,5.6){\vector(-1,0){1.3}} 
\put(6.4,5.8){\line(-1,1){1.3}}    
\put(1.2,5.6){\circle*{0.2}}
\put(1.2,6.1){$X_3$}
\put(1.2,3.8){\circle*{0.2}}
\put(1.2,4.3){$X_1$}
\put(1.4,3.8){\line(1,0){1.3}}
\put(3,3.8){\circle*{0.2}} 
\put(2.4,4.3){$X_2$}
\put(3.2,3.8){\vector(1,0){1.3}} 
\put(4.7,3.8){\circle*{0.2}} 
\put(4.4,4.3){$X_6$}
\put(4.9,3.6){\line(1,-1){1.3}}    
\end{picture}
\end{center}
\caption{A partially oriented graph due to Principle II(nodes $X_4,X_9$)}
\label{abbdrei}
\end{figure}
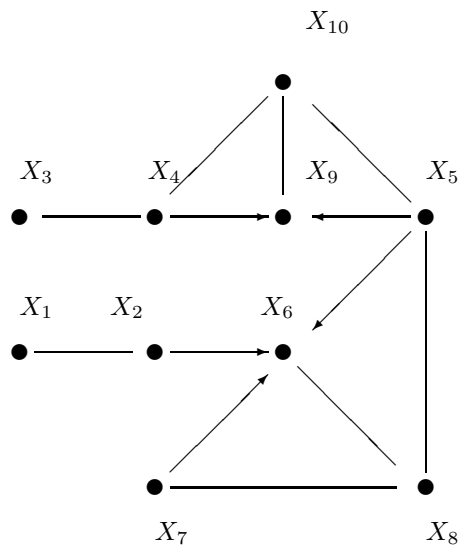
} 
%
\newcommand{\AbbVier}
{
\begin{figure}
\begin{center}
\begin{picture}(8,8.3)
\put(3,2){\circle*{0.2}} 
\put(3,1.3){$X_7$}
\put(3.2,2){\line(1,0){3.0}}
\put(3.2,2.2){\vector(1, 1){1.3}}    
\put(6.6,2){\circle*{0.2}}
\put(6.6,1.3){$X_8$}
\put(3,5.6){\circle*{0.2}}
\put(2.8,5.6){\line(-1,0){1.3}}  
\put(2.9,6.1){$X_4$} 
\put(3.2,5.6){\vector(1,0){1.3}}   
\put(4.7,7.4){\circle*{0.2}} 
\put(5,8.1){$X_{10}$} 
\put(4.5,7.2){\line(-1,-1){1.3}}    
\put(4.7,7.2){\line(0,-1){1.3}}   
\put(4.7,5.6){\circle*{0.2}} 
\put(5,6.1){$X_9$}
\put(6.6,5.6){\circle*{0.2}}
\put(6.6,6.1){$X_5$}
\put(6.4,5.4){\vector(-1, -1){1.3}}    
\put(6.6,3.4){\line(0,-1){1.2}}    
\put(6.6,5.4){\line(0,-1){3.1}}   
\put(6.4,5.6){\vector(-1,0){1.3}} 
\put(6.4,5.8){\line(-1,1){1.3}}    
\put(1.2,5.6){\circle*{0.2}}
\put(1.2,6.1){$X_3$}
\put(1.2,3.8){\circle*{0.2}}
\put(1.2,4.3){$X_1$}
\put(1.4,3.8){\line(1,0){1.3}}
\put(3,3.8){\circle*{0.2}} 
\put(2.4,4.3){$X_2$}
\put(3.2,3.8){\vector(1,0){1.3}} 
\put(4.7,3.8){\circle*{0.2}} 
\put(4.4,4.3){$X_6$}
\put(4.9,3.6){\vector(1,-1){1.3}}    
\end{picture}
\end{center}
\caption{A partially oriented graph due to Principle II$^c$ 
(arrow  $(X_6,X_8)$)}
\label{abbvier}
\end{figure}
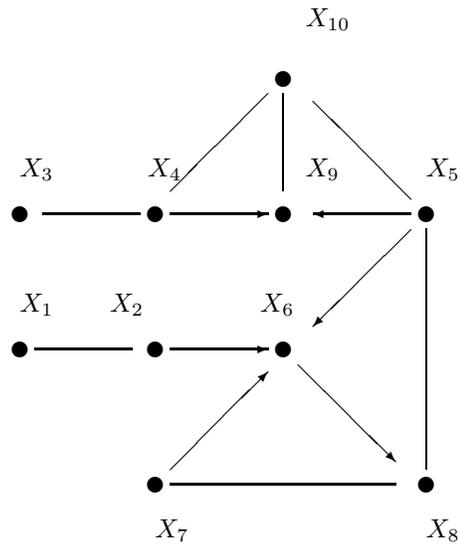
} 
%
%
\newcommand{\AbbFuenf}
{
\begin{figure}
\begin{center}
\begin{picture}(8,8.3)
\put(3,2){\circle*{0.2}} 
\put(3,1.3){$X_7$}
\put(3.2,2){\vector(1,0){1.3}} 
\put(4.6,2){\line(1,0){1.7}}   
\put(3.2,2.2){\vector(1, 1){1.3}}    
\put(6.6,2){\circle*{0.2}}
\put(6.6,1.3){$X_8$}
\put(3,5.6){\circle*{0.2}}
\put(2.8,5.6){\line(-1,0){1.3}}  
\put(2.9,6.1){$X_4$} 
\put(3.2,5.6){\vector(1,0){1.3}}   
\put(4.7,7.4){\circle*{0.2}} 
\put(5,8.1){$X_{10}$} 
\put(4.5,7.2){\line(-1,-1){1.3}}    
\put(4.7,7.2){\line(0,-1){1.3}}   
\put(4.7,5.6){\circle*{0.2}} 
\put(5,6.1){$X_9$}
\put(6.6,5.6){\circle*{0.2}}
\put(6.6,6.1){$X_5$}
\put(6.4,5.4){\vector(-1, -1){1.3}}    
\put(6.6,5.4){\vector(0,-1){1.9}}  
\put(6.6,5.4){\line(0,-1){3.1}}   
\put(6.4,5.6){\vector(-1,0){1.3}} 
\put(6.4,5.8){\line(-1,1){1.3}}    
\put(1.2,5.6){\circle*{0.2}}
\put(1.2,6.1){$X_3$}
\put(1.2,3.8){\circle*{0.2}}
\put(1.2,4.3){$X_1$}
\put(1.4,3.8){\line(1,0){1.3}}
\put(3,3.8){\circle*{0.2}} 
\put(2.4,4.3){$X_2$}
\put(3.2,3.8){\vector(1,0){1.3}} 
\put(4.7,3.8){\circle*{0.2}} 
\put(4.4,4.3){$X_6$}
\put(4.9,3.6){\vector(1,-1){1.3}}    
\end{picture}
\end{center}
\caption{A partially oriented graph due to Principle IV 
(arrows  $(X_7,X_8)$, $(X_5,X_8)$)}
\label{abbfuenf}
\end{figure}
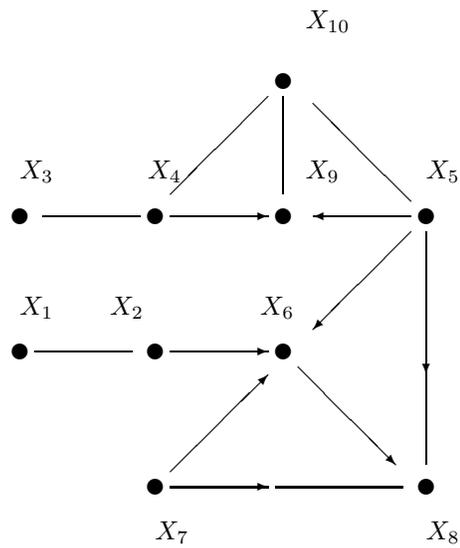
} 
%
%
\newcommand{\AbbSechs}
{
\begin{figure}
\begin{center}
\begin{picture}(8,8.3)
\put(3,2){\circle*{0.2}} 
\put(3,1.3){$X_7$}
\put(3.2,2){\vector(1,0){1.3}} 
\put(4.6,2){\line(1,0){1.7}}   
\put(3.2,2.2){\vector(1, 1){1.3}}    
\put(6.6,2){\circle*{0.2}}
\put(6.6,1.3){$X_8$}
\put(3,5.6){\circle*{0.2}}
\put(2.8,5.6){\line(-1,0){1.3}}  
\put(2.9,6.1){$X_4$} 
\put(3.2,5.6){\vector(1,0){1.3}}   
\put(4.7,7.4){\circle*{0.2}} 
\put(5,8.1){$X_{10}$} 
\put(4.5,7.2){\line(-1,-1){1.3}}    
\put(4.7,7.2){\vector(0,-1){0.9}}  
\put(4.7,6.2){\line(0,-1){0.4}}    
\put(4.7,5.6){\circle*{0.2}} 
\put(5,6.1){$X_9$}
\put(6.6,5.6){\circle*{0.2}}
\put(6.6,6.1){$X_5$}
\put(6.4,5.4){\vector(-1, -1){1.3}}    
\put(6.6,5.4){\vector(0,-1){1.9}}  
\put(6.6,5.4){\line(0,-1){3.1}}   
\put(6.4,5.6){\vector(-1,0){1.3}} 
\put(6.4,5.8){\line(-1,1){1.3}}    
\put(1.2,5.6){\circle*{0.2}}
\put(1.2,6.1){$X_3$}
\put(1.2,3.8){\circle*{0.2}}
\put(1.2,4.3){$X_1$}
\put(1.4,3.8){\line(1,0){1.3}}
\put(3,3.8){\circle*{0.2}} 
\put(2.4,4.3){$X_2$}
\put(3.2,3.8){\vector(1,0){1.3}} 
\put(4.7,3.8){\circle*{0.2}} 
\put(4.4,4.3){$X_6$}
\put(4.9,3.6){\vector(1,-1){1.3}}    
\end{picture}
\end{center}
\caption{A partially oriented graph due to Principle V 
(arrow  $(X_10,X_9)$). Nodes $X_1$,  $X_3$, $X_8$, $X_9$ are legitimately 
removable.} \label{abbsechs}
\end{figure}
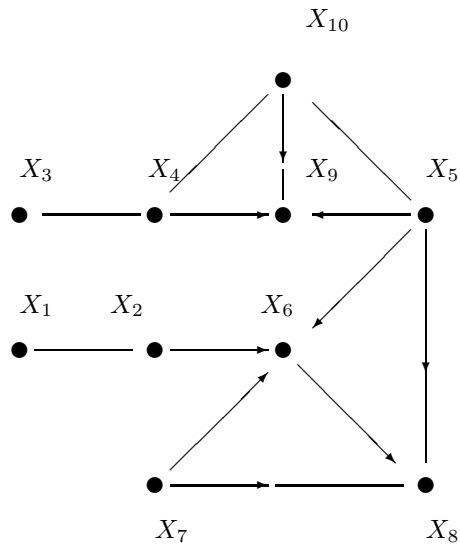
} 
%
%
%
\newcommand{\AbbSieben}
{
\begin{figure}
\begin{center}
\begin{picture}(8,8.3)
\put(3,2){\circle*{0.2}} 
\put(3,1.3){$X_7$}
\put(3.2,2.2){\vector(1, 1){1.3}}    
\put(3,5.6){\circle*{0.2}}
\put(2.9,6.1){$X_4$} 
\put(4.7,7.4){\circle*{0.2}} 
\put(5,8.1){$X_{10}$} 
\put(4.5,7.2){\line(-1,-1){1.3}}    
\put(6.6,5.6){\circle*{0.2}}
\put(6.6,6.1){$X_5$}
\put(6.4,5.4){\vector(-1, -1){1.3}}    
\put(6.4,5.8){\line(-1,1){1.3}}    
\put(1.2,3.8){\circle*{0.2}}
\put(1.2,4.3){$X_1$}
\put(1.4,3.8){\line(1,0){1.3}}
\put(3,3.8){\circle*{0.2}} 
\put(2.4,4.3){$X_2$}
\put(3.2,3.8){\vector(1,0){1.3}} 
\put(4.7,3.8){\circle*{0.2}} 
\put(4.4,4.3){$X_6$}
\end{picture}
\end{center}
\caption{After legitimate removal of nodes $X_3$, $X_8$ and $X_9$.
(The arrow $(X_3,X_4)$ was inforced).
 Nodes $X_1$,  $X_4$, $X_6$ are legitimately 
removable.} \label{abbsieben}
\end{figure}
} 
%
%
\newcommand{\AbbAcht}
{
\begin{figure}
\begin{center}
\begin{picture}(8,8.3)
\put(1.0,2.5){\circle*{0.2}} 
\put(1.2,2.7){$X_{11}$}
\put(1.2,2.5){\vector(1,0){1.5}} 
\put(1.2,2.7){\vector(1,1){1.5}} 
\put(1.2,2.3){\vector(1,-1){1.5}} 
\put(1.0,4.5){\circle*{0.2}} 
\put(1.2,4.7){$X_{16}$}
\put(3.0,0.5){\circle*{0.2}} 
\put(3.2,0.3){$X_{13}$}
\put(3.2,0.5){\vector(1,0){1.5}} 

\put(3.0,2.5){\circle*{0.2}} 
\put(3.2,2.7){$X_{14}$}
\put(3.0,2.7){\line(0,1){1.5}} 
\put(3.0,2.3){\line(0,-1){1.5}} 
\put(3.0,4.5){\circle*{0.2}} 
\put(3.2,4.7){$X_{15}$}
\put(2.8,4.5){\vector(-1,0){1.5}} 
\put(5.0,2.5){\circle*{0.2}} 
\put(5.2,2.7){$X_{12}$}
\put(4.8,2.5){\vector(-1,0){1.5}} 
\put(4.8,2.7){\vector(-1,1){1.5}} 
\put(4.8,2.3){\vector(-1,-1){1.5}} 
\put(5.2,0.7){$X_{17}$}
\end{picture}
\end{center}
\caption{A partially oriented graph. Active trails and active p-trails.
} \label{abbacht}
\end{figure}
} 
%
%
\newcommand{\AbbNeun}
{
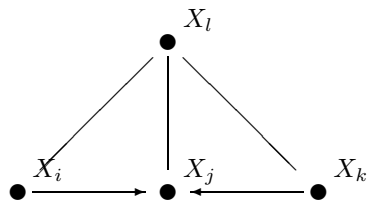
\begin{figure}
\begin{center}
\begin{picture}(8,3.1)
\put(1.0,0.5){\circle*{0.2}} 
\put(1.2,0.5){\vector( 1,0){1.5}} 
\put(1.2,0.7){$X_i$}
\put(3.0,0.5){\circle*{0.2}} 
\put(3.2,0.7){$X_j$}
\put(5.0,0.5){\circle*{0.2}} 
\put(5.2,0.7){$X_k$}
\put(4.8,0.5){\vector(-1,0){1.5}} 
\put(3.0,2.5){\circle*{0.2}} 
\put(3.2,2.7){$X_l$}
\put(2.8,2.3){\line(-1,-1){1.5}} 
\put(3.0,2.3){\line( 0,-1){1.5}} 
\put(3.2,2.3){\line( 1,-1){1.5}} 
\end{picture}
\end{center}
\caption{Visualisation to the Proof of the Theorem on Principle V         
} \label{abbneun}
\end{figure}
} 
%
%
\newcommand{\AbbZehn}
{
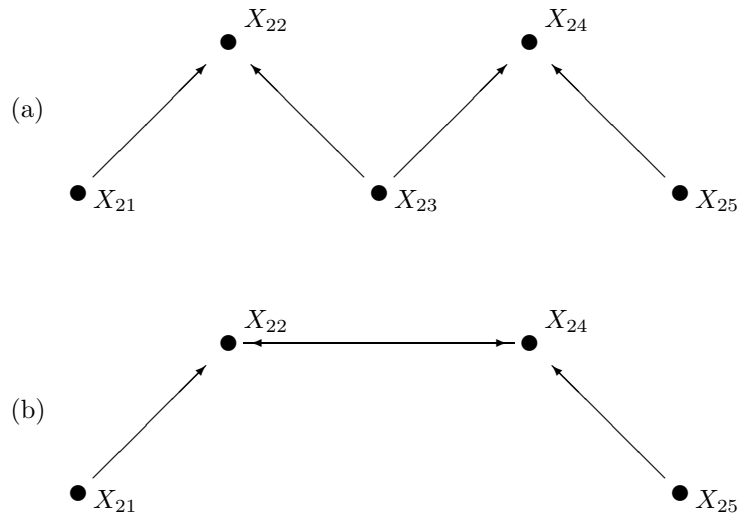
\begin{figure}
\begin{center}
\begin{picture}(9.5,8.3)
\put(0.1,5.5){(a)}
\put(1.0,4.5){\circle*{0.2}} 
\put(1.2,4.3){$X_{21}$}
\put(1.2,4.7){\vector( 1, 1){1.5}} 
\put(5.0,4.5){\circle*{0.2}} 
\put(5.2,4.3){$X_{23}$}
\put(4.8,4.7){\vector(-1, 1){1.5}} 
\put(5.2,4.7){\vector( 1, 1){1.5}} 
\put(9.0,4.5){\circle*{0.2}} 
\put(9.2,4.3){$X_{25}$}
\put(8.8,4.7){\vector(-1, 1){1.5}} 
\put(3.0,6.5){\circle*{0.2}} 
\put(3.2,6.7){$X_{22}$}
\put(7.0,6.5){\circle*{0.2}} 
\put(7.2,6.7){$X_{24}$}
\put(0.1,1.5){(b)}
\put(1.0,0.5){\circle*{0.2}} 
\put(1.2,0.3){$X_{21}$}
\put(1.2,0.7){\vector( 1, 1){1.5}} 
\put(9.0,0.5){\circle*{0.2}} 
\put(9.2,0.3){$X_{25}$}
\put(8.8,0.7){\vector(-1, 1){1.5}} 
\put(3.0,2.5){\circle*{0.2}} 
\put(3.2,2.7){$X_{22}$}
\put(3.2,2.5){\vector( 1,0 ){3.5}} 
\put(7.0,2.5){\circle*{0.2}} 
\put(7.2,2.7){$X_{24}$}
\put(6.8,2.5){\vector(-1,0 ){3.5}} 
\end{picture}
\end{center}
\caption{An effect of variable hiding 
(a) an original dag (b) after "hiding" $X_{23}$  - double orientation of an 
edge (connecting $X_{22}$ with $X_{24}$)} \label{abbzehn}
\end{figure}
} 
%
%
\newcommand{\AbbElf }
{
\begin{figure}
\begin{center}
\begin{picture}(8,8.3)
\put(0.1,2.7){(a)}
\put(1.0,2.5){\circle*{0.2}} 
\put(1.2,2.7){$X_{31}$}
\put(1.2,2.3){\vector(1,-1){1.5}} 
\put(1.0,4.5){\circle*{0.2}} 
\put(1.2,4.7){$X_{36}$}
\put(1.2,4.5){\vector( 1,0){1.5}} 
\put(3.0,0.5){\circle*{0.2}} 
\put(3.1,0.1){$X_{33}$}

\put(1.0,0.5){\circle*{0.2}} 
\put(1.2,0.7){$X_{34}$}
\put(1.0,0.7){\vector(0,1){1.5}} 
\put(3.0,4.5){\circle*{0.2}} 
\put(3.2,4.7){$X_{35}$}
\put(3.2,4.7){\vector(-1,-1){1.5}} 
\put(5.0,2.5){\circle*{0.2}} 
\put(5.2,2.7){$X_{32}$}
\put(4.8,2.7){\vector(-1,1){1.5}} 
\put(4.8,2.3){\vector(-1,-1){1.5}} 
\put(5.0,2.3){\vector( 0,-1){1.5}} 
\put(5.0,0.5){\circle*{0.2}} 
\put(5.2,0.7){$X_{37}$}
\put(4.8,0.5){\vector(-1,0){1.5}} 
\put(6.1,2.7){(b)}
\put(7.0,2.5){\circle*{0.2}} 
\put(7.2,2.7){$X_{31}$}
\put(7.2,2.3){\vector(1,-1){1.5}} 
\put(7.0,4.5){\circle*{0.2}} 
\put(7.2,4.7){$X_{36}$}
\put(7.2,4.5){\vector( 1,0){1.5}} 
\put(9.0,0.5){\circle*{0.2}} 
\put(9.1,0.1){$X_{33}$}
\put(9.0,0.7){\vector(0,1){3.5}} 

\put(7.0,0.5){\circle*{0.2}} 
\put(7.2,0.7){$X_{34}$}
\put(7.0,0.7){\vector(0,1){1.5}} 
\put(9.0,4.5){\circle*{0.2}} 
\put(9.2,4.7){$X_{35}$}
\put(9.2,4.7){\vector(-1,-1){1.5}} 
\put(11.0,0.5){\circle*{0.2}} 
\put(11.2,0.7){$X_{37}$}
\put(10.8,0.5){\vector(-1,0){1.5}} 
\put(10.8,0.7){\vector(-1,2){1.7}} 

\end{picture}
\end{center}
\caption{An effect of variable hiding (a) an original dag (b)
hiding $X_{33}$ leads to anoriented 
cycle } \label{abbelf}
\end{figure}
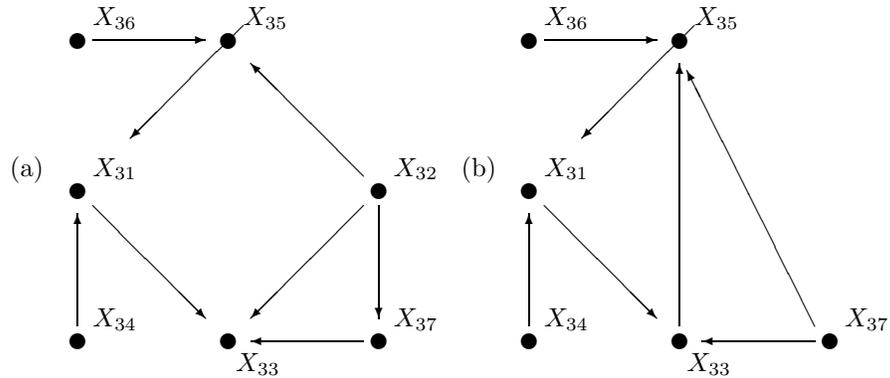
} 
\newcommand{\AbbZwoelf}
{
\begin{figure}
\begin{center}
\begin{picture}(6,8.3)
\put(0,8){a)}
\put(2,2){\circle*{0.2}} 
\put(2,1.3){$X_7$}
\put(2.2,2){\vector(1,0){1.3}}
\put(3.6,2){\line(1,0){1.7}}
\put(2.2,2.2){\vector(1, 1){1.3}}    
\put(5.6,2){\circle*{0.2}}
\put(5.6,1.3){$X_8$}
\put(2,5.6){\circle*{0.2}}
\put(1.9,6.1){$X_4$} 
\put(1.8,5.6){\vector(-1,0){0.9}}  
\put(0.8,5.6){\line(-1,0){0.4}}    
\put(2.2,5.6){\vector(1,0){1.3}}
\put(3.7,7.4){\circle*{0.2}} 
\put(4,8.1){$X_{10}$} 
\put(3.5,7.2){\vector(-1,-1){1.3}}    
\put(3.7,7.2){\vector(0,-1){0.9}}
\put(3.7,6.2){\line(0,-1){0.4}}
\put(3.7,5.6){\circle*{0.2}} 
\put(4,6.1){$X_9$}
\put(5.6,5.6){\circle*{0.2}}
\put(5.6,6.1){$X_5$}
\put(5.4,5.4){\vector(-1, -1){1.3}}    
\put(5.6,5.4){\vector(0,-1){1.9}}
\put(5.6,3.4){\line(0,-1){1.2}}
\put(5.4,5.6){\vector(-1,0){1.3}}
\put(5.4,5.8){\vector(-1,1){1.3}}    
\put(0.2,5.6){\circle*{0.2}}
\put(0.2,6.1){$X_3$}
\put(0.2,3.8){\circle*{0.2}}
\put(0.2,4.3){$X_1$}
\put(0.4,3.8){\vector(1,0){0.9}}
\put(1.4,3.8){\line(1,0){0.4}}
\put(2,3.8){\circle*{0.2}} 
\put(1.4,4.3){$X_2$}
\put(2.2,3.8){\vector(1,0){1.3}}
\put(3.7,3.8){\circle*{0.2}} 
\put(3.4,4.3){$X_6$}
\put(3.5,4.0){\vector(-1,1){1.3}}    
\put(3.9,3.6){\vector(1,-1){1.3}}    
\end{picture}
\begin{picture}(6,8.3)
\put(1,8){b)}
\put(3,2){\circle*{0.2}} 
\put(3,1.3){$X_7$}
\put(3.2,2){\vector(1,0){1.3}}
\put(4.6,2){\line(1,0){1.7}}
\put(3.2,2.2){\vector(1, 1){1.3}}    
\put(6.6,2){\circle*{0.2}}
\put(6.6,1.3){$X_8$}
\put(3,5.6){\circle*{0.2}}
\put(2.9,6.1){$X_4$} 
\put(2.8,5.6){\vector(-1,0){0.9}}  
\put(1.8,5.6){\line(-1,0){0.4}}    
\put(3.2,5.6){\vector(1,0){1.3}}
\put(4.7,7.4){\circle*{0.2}} 
\put(5,8.1){$X_{10}$} 
\put(4.5,7.2){\vector(-1,-1){1.3}}    
\put(4.7,7.2){\vector(0,-1){0.9}}
\put(4.7,6.2){\line(0,-1){0.4}}
\put(4.7,5.6){\circle*{0.2}} 
\put(5,6.1){$X_9$}
\put(6.6,5.6){\circle*{0.2}}
\put(6.6,6.1){$X_5$}
\put(6.4,5.4){\vector(-1, -1){1.3}}    
\put(6.6,5.4){\vector(0,-1){1.9}}
\put(6.6,3.4){\line(0,-1){1.2}}
\put(6.4,5.6){\vector(-1,0){1.3}}
\put(6.4,5.8){\vector(-1,1){1.3}}    
\put(1.2,5.6){\circle*{0.2}}
\put(1.2,6.1){$X_3$}
\put(1.2,3.8){\circle*{0.2}}
\put(1.2,4.3){$X_1$}
\put(1.4,3.8){\vector(1,0){0.9}}
\put(2.4,3.8){\line(1,0){0.4}}
\put(3,3.8){\circle*{0.2}} 
\put(2.4,4.3){$X_2$}
\put(3.2,3.8){\vector(1,0){1.3}}
\put(4.7,3.8){\circle*{0.2}} 
\put(4.4,4.3){$X_6$}
\end{picture}
\end{center}
\caption{Both dags a) and b) represent the same independence information
if 
 $X_6$ is in the d-separating set. In b) directed edges $(X_6,X_3)$ and  
$(X_6,X_8)$ have been removed. }  
\label{abbzwoelf}
\end{figure}
} 
\newcommand{\AbbDreizehn}
{
\begin{figure}
\begin{center}
\begin{picture}(6,8.3)
\put(0,8){a)}
\put(2,2){\circle*{0.2}} 
\put(2,1.3){$X_7$}
\put(2.2,2){\vector(1,0){1.3}}
\put(3.6,2){\line(1,0){1.7}}
\put(2.2,2.2){\vector(1, 1){1.3}}    
\put(5.6,2){\circle*{0.2}}
\put(5.6,1.3){$X_8$}
\put(2,5.6){\circle*{0.2}}
\put(1.9,6.1){$X_4$} 
\put(1.8,5.6){\vector(-1,0){0.9}}  
\put(0.8,5.6){\line(-1,0){0.4}}    
\put(2.2,5.6){\vector(1,0){1.3}}
\put(3.7,7.4){\circle*{0.2}} 
\put(4,8.1){$X_{10}$} 
\put(3.5,7.2){\vector(-1,-1){1.3}}    
\put(3.7,7.2){\vector(0,-1){0.9}}
\put(3.7,6.2){\line(0,-1){0.4}}
\put(3.7,5.6){\circle*{0.2}} 
\put(4,6.1){$X_9$}
\put(5.6,5.6){\circle*{0.2}}
\put(5.6,6.1){$X_5$}
\put(5.4,5.4){\vector(-1, -1){1.3}}    
\put(5.6,5.4){\vector(0,-1){1.9}}
\put(5.6,3.4){\line(0,-1){1.2}}
\put(5.4,5.6){\vector(-1,0){1.3}}
\put(5.4,5.8){\vector(-1,1){1.3}}    
\put(0.2,5.6){\circle*{0.2}}
\put(0.2,6.1){$X_3$}
\put(0.2,3.8){\circle*{0.2}}
\put(0.2,4.3){$X_1$}
\put(0.4,3.8){\vector(1,0){0.9}}
\put(1.4,3.8){\line(1,0){0.4}}
\put(2,3.8){\circle*{0.2}} 
\put(1.4,4.3){$X_2$}
\put(2.2,3.8){\vector(1,0){1.3}}
\put(3.7,3.8){\circle*{0.2}} 
\put(3.4,4.3){$X_6$}
\end{picture}
\begin{picture}(6,8.3)
\put(1,8){b)}
\put(3,2){\circle*{0.2}} 
\put(3,1.3){$X_7$}
\put(3.2,2){\vector(1,0){1.3}}
\put(4.6,2){\line(1,0){1.7}}
\put(3.2,2.2){\vector(1, 1){1.3}}    
\put(3  ,2.2){\vector(0, 1){1.3}}    
\put(6.6,2){\circle*{0.2}}
\put(6.6,1.3){$X_8$}
\put(3,5.6){\circle*{0.2}}
\put(2.9,6.1){$X_4$} 
\put(2.8,5.6){\vector(-1,0){0.9}}  
\put(1.8,5.6){\line(-1,0){0.4}}    
\put(3.2,5.6){\vector(1,0){1.3}}
\put(4.7,7.4){\circle*{0.2}} 
\put(5,8.1){$X_{10}$} 
\put(4.5,7.2){\vector(-1,-1){1.3}}    
\put(4.7,7.2){\vector(0,-1){0.9}}
\put(4.7,6.2){\line(0,-1){0.4}}
\put(4.7,5.6){\circle*{0.2}} 
\put(5,6.1){$X_9$}
\put(6.6,5.6){\circle*{0.2}}
\put(6.6,6.1){$X_5$}
\put(6.4,5.4){\vector(-1, -1){1.3}}    
\put(6.2,5.4){\vector(-2, -1){2.9}}    
\put(6.6,5.4){\vector(0,-1){1.9}}
\put(6.6,3.4){\line(0,-1){1.2}}
\put(6.4,5.6){\vector(-1,0){1.3}}
\put(6.4,5.8){\vector(-1,1){1.3}}    
\put(1.2,5.6){\circle*{0.2}}
\put(1.2,6.1){$X_3$}
\put(1.2,3.8){\circle*{0.2}}
\put(1.2,4.3){$X_1$}
\put(1.4,3.8){\vector(1,0){0.9}}
\put(2.4,3.8){\line(1,0){0.4}}
\put(1.4,4.0){\line(2,1){1.6}} 
\put(3  ,4.8){\vector(2,-1){1.5}} 
\put(3,3.8){\circle*{0.2}} 
\put(2.4,4.3){$X_2$}
\put(4.7,3.8){\circle*{0.2}} 
\put(4.4,4.3){$X_6$}
\put(4.5,3.8){\vector(-1,0){1.3}} 
\end{picture}
\end{center}
\caption{Both dags a) and b) represent the same independence information
if 
 $X_6$ is in the d-separating set. In b)
the direction of edge $(X_2,X_6)$ has been reversed and edges
 $(X_1,X_6)$,     
$(X_5,X_2)$  and $(X_7,X_2)$ 
have been added
}  
\label{abbdreizehn}
\end{figure}
} 
%
%




\begin{abstract}
Shenoy and Shafer \cite{Shenoy:90} demonstrated that both for Dempster-Shafer 
Theory and probability theory there exists a possibility to calculate 
efficiently marginals of joint belief distributions (by so-called local 
computations) provided that the joint distribution can be decomposed 
(factorized) into a belief network. A number of algorithms exists for 
decomposition of probabilistic joint belief distribution into a bayesian 
(belief) network from data. For example 
 Spirtes, Glymour and Scheines \cite{Spirtes:90b} formulated a Conjecture
that a direct dependence test and a head-to-head meeting test
would suffice to construe bayesian network  from data             
in such a way that 
Pearl's concept of d-separation \cite{Geiger:90} applies. \\
 This paper is intended to transfer 
 Spirtes, Glymour and Scheines \cite{Spirtes:90b} approach onto the ground of 
the Dempster-Shafer Theory (DST). For this purpose, a frequentionistic 
interpretation of the DST developed in \cite{Klopotek:93b} is exploited. A 
special notion of conditionality for DST is introduced and demonstrated to 
behave with respect to Pearl's d-separation  \cite{Geiger:90} much the same 
way as conditional probability (though some differences like non-uniqueness 
are evident). Based on this, an algorithm analogous to that from  
\cite{Spirtes:90b} is developed.
 The notion of a partially oriented graph (pog) is introduced and 
within 
this graph the notion of p-d-separation is defined. If  direct 
 dependence test and  head-to-head meeting test are used to orient the pog 
then 
its p-d-separation is shown to be equivalent to the Pearl's d-separation for 
any compatible dag.

\end{abstract}

{\bf Keywords:} Dempster-Shafer belief network; learning from data; 
conditional beliefs; d-separation; partially oriented belief networks.

\font\gh=eufm10 scaled \magstep1
\newcommand{\Prob}[2]{{ {\mbox{\gh Prob} ^{#2(#1)}} 
                         \atop {_{#1}} 
                     }}
\section{Introduction}

Many researchers consider the theory of evidence  a 
proper tool for representation of uncertainty. 
It has been developed by Dempster \cite{Dempster:67} and Shafer 
\cite{Shafer:76} and possesses several interesting properties. 
one of them is that if we are able to factorize a joint belief distribution
 into a hyper-graph structure (that is to represent it as combination of 
simpler 
belief functions) then there exists
the possibility of local computations of marginals
of the joint belief distribution 
as well as conditionals of some variables on various events
without actually calculating this joint 
belief distribution \cite{Shenoy:90}. This is of tremendous importance if 
we imagine how much space in computer memory would be required to represent a 
joint belief distribution in, say, 20 variables. The Shenoy-Shafer theory of 
local computations makes marginalizations       and calculation of 
conditionals for 
belief functions more feasible. 

Actually to exploit this fine property we need a tool for factorizing the 
joint belief distribution into such factors. From the experience with 
probability distributions it is known that they may be represented in form of 
bayesian networks, that is  directed  acyclic  graphs  reflecting 
dependencies 
between variables. These directed acyclic graphs (dag) may be transformed 
directly 
to 
hypergraphs being precisely 
frameworks for
factoriozatiions required by the 
Shenoy-Shafer 
theory of local computations. 
A number of techniques of decomposition of a joint probability distribution 
into  bayesian networks have been developed.
In this paper we want to find an analogous decomposition of a joint belief 
distribution following the outlines of Spirtes et al approach 
\cite{Spirtes:90b}, which has been developed for probability distributions.

In Section 2 we will briefly recall basic definitions of the DS theory of 
evidence. Section 3 will recall the Shenoy-Shafer requirements for local 
computations of a joint distribution. Section 4 is devoted to selection of 
proper conditional belief definition. Section 5 introduces my own sense 
of conditionality. Section 6 develops the algorithm. Section 7 contains some 
conclusions. 

\section{Formal  Definition  of  the  Dempster-Shafer  Theory   of 
Evidence}

Let us make the remark that if an object is described by a set of 
discrete attributes (features, variables) $X_1,X_2,...,X_n$ taking  values  
from their respective domains $\Xi_1,\Xi_2,...,\Xi_n$  then we can 
think of it  as 
being described by a complex attribute $X$  having  vector  values, 
that is the domain $\Xi$ of X is equal:
$$\Xi=\{(x_1,x_2,...,x_n) | x_i \in \Xi_i \forall i=1,...,n\}$$.
However sometimes we will treat $X$ as a set of attributes applying 
set-theoretic operations $\cap, \cup, -$ understanding, that there exists some 
natural ordering among the attributes constituting $X$ so that any non-empty 
"subset" $Y$ of $X$  $Y\subseteq X$ is a Cartesian product of its 
"components": $Y=X_{k_1}\times X_{k_2} \times \dots\times X_{k_m}$, where 
indices $k_1,k_2,...,k_m$ are an increasing subsequence from the sequence 
$1,2,...,n$.

So in definitions below  let us assume that we  are  talking 
of objects described by a single attribute X  taking  its  values 
from the domain $\Xi$. We say that $\Xi$, the domain of X
 is our space of discernment 
spanned by the attribute X. We shall also briefly say that X is our
space of discernment instead.

The function m (Mass Function, or basic probability assignment function 
bpa) is defined as 
\begin{df} \label{mDef}
The Pseudo-Mass Function in the sense of the DS-Theory is defined as
m:$2^\Xi  \rightarrow [-1,1]$
with
$$  \sum_{A \in 2^\Xi } |m(A)|=1 $$
$$  m(\{\emptyset\}=0 $$
$$\forall_{A \in 2^\Xi} \quad  0 \leq  \sum_{A \subseteq B} m(B)$$
where $|.|$ means the absolute value operator.
If also
$$  \forall_{ A \in 2^\Xi} \quad m(A) \geq 0 $$
holds, than we talk of (intrinsic) Mass Function.
\end{df}
\begin{df}
 Whenever $m(A) > 0$, we say that A is the focal  point  of  the 
Bel-Function.
\end{df}

For the purpose of  this  paper  we  define  the  Bel-function  as 
follows.

\begin{df} \label{BelDef}
  The Pseudo-Belief Function in the sense of the DS-Theory is defined 
as Bel:$2^\Xi \rightarrow [-1,1]$ with 
 $$Bel(A) = \sum_{B \subseteq A} m(B)$$
for any nonempty $A \in 2^\Xi$ where m(A) is a Pseudo-Mass Function in the 
sense of the 
 DS-Theory (see \rDef{mDef} above). 
If m(A) is an (intrinsic) Mass Function, then Bel is called the (intrinsic) 
Belief Function (the word "intrinsic" is omitted subsequently).
\end{df}

Let  us  also  introduce  the  Pl-Function   (Plausibility)   for 
non-empty sets A as:
\begin{df} The Pseudo-Plausibility Function in the sense of the DS-Theory is 
defined as 
Pl:$2^\Xi \rightarrow [-1,1]$ with 
$$\forall_{A \in 2^\Xi} \  Pl(A) = 1-Bel(\Xi-A )$$
If Bel is a Belief Function then Pl is a Plausibility Function.
\end{df}

For completeness let us recall also the Q-Function of the DS-Theory.

\begin{df} The Q-Function in the sense of the DS-Theory is 
defined as Q:$2^\Xi \rightarrow [0,1]$ with 

 $$\forall_{A \in 2^\Xi} \quad Q(A) = \sum_{A \subseteq B} m(B)$$
where m(A) is a Pseudo-Mass Function in the sense of the DS-Theory 
(see \rDef{mDef} above and notice the difference to \rDef{BelDef} in 
that the sum is taken over supersets, not subsets of A). 
\end{df}

Please pay attention to the fact, that domains of values of Mass Function, 
Belief Function and Plausibility Function are always $[0,1]$, whereas those 
of their Pseudo-Counterparts are $[-1,1]$.  However, the domain of the 
Q-Function is always $[0,1]$ independly of whether m is a Pseudo-Mass or 
(intrinsic) Mass Function.

Please pay attention to the fact that knowing one of the functions m, Bel, Pl 
or Q suffices to derive any other of them.

Beside the above  definition  a  characteristic  feature  of  the 
DS-Theory is the so-called DS-rule of combination of  (independent) 
evidence:
\begin{df}

   Let   $Bel_{E_1}$    and    $Bel_{E_2}$     represent 
independent information over the same space of discernment. Then:
    $$Bel_{E_1,E_2}=Bel_{E_1} \oplus Bel_{E_2}$$ 
calculated as:
$$m_{E_1,E_2}(A)=c \cdot  \sum_{B,C; A= B \cap C} m_{E_1}(B) \cdot 
m_{E_2}(C)$$ (c - normalizing constant, 
$c = ( \sum_{B,C; B \cap C \ne \emptyset} 
m_{E_1}(B)\cdot m_{E_2}(C))^{-1} )$ represents the Combined 
Pseudo-Belief-Function  of 
Two Independent Pseudo-Beliefs
\end{df}

Let us also introduce the marginalization and extension operations:
first for sets.

\begin{df}
 Let $X=X_1 \times X_2 \times \dots \times X_n$ and 
$\Xi = \Xi_1 \times \Xi_2 \times \dots \times \Xi_n$. Let A be a 
set $A \in 2^\Xi$. Let $Y=X_{i_1} \times X_{i_2} \times \dots \times 
X_{i_k}$, where indices $\{i_1,\dots,i_k\}$ are all distinct and are 
 subset of \{1,...,n\}.
The set B is called  the {\em projection (marginalization)}
 of the set A onto the (sub)space Y (denoted
$B=A ^{\downarrow Y}$) iff for every element $(v_1,\dots,v_n) \in A$
the element  $(v_{i_1}, v_{i_2}, \dots , v_{i_k})$ belongs to B. \\
We shall say also say that A is an {\em extension} of B.\\
\end{df}

We shall distinguish one special extension: the empty extension.\\

\begin{df}
Let $B \subseteq \Xi_1$. Let $A \subseteq \Xi_1 \times \Xi_2$ such that
$A=B \times \Xi_2$. Then we say that 
A is the {\em empty extension} of B, denoted $A = B ^{\uparrow X}$..
\end{df}

Now let us define the marginalization and extension for Bel-Functions:

\begin{df}
Let $X =X_1 \times X_2$ be our space of discernment for which the m, and its  
Bel, Pl and Q functions are defined. \\
The m function marginalized (projected)  
onto the subspace $X_1$, denoted as $m ^{\downarrow X_1}$ is defined as:
$$\forall_{B;B \subseteq \Xi_1} \quad
m ^{\downarrow X_1}(B) =
\sum_{A;B= A ^{\downarrow X_1}} m(A)$$.
The functions  $Bel ^{\downarrow X_1}$, $Pl ^{\downarrow X_1}$  and
 $Q ^{\downarrow X_1}$
are 
defined accordingly to Bel, Pl and Q definitions above with respect to 
 $m ^{\downarrow X_1}$ as their (pseudo-)mass function.\\
\end{df}

\begin{df}
Let $X_1$ be our space of discernment for which the m, and its  
Bel, Pl and Q  functions are defined. \\
The m function empty-extended 
onto the superspace $X=X_1 \times X_2$, denoted as $m ^{\uparrow X}$ is 
defined as:
$$\forall_{A;A \subseteq \Xi, A = (A ^{\downarrow X_1})^{\uparrow X}}
\quad
  m ^{\uparrow X}(A) = m(A ^{\downarrow X_1})$$
and 
$$\forall_{A;A \subseteq \Xi, A \neq (A ^{\downarrow X_1})^{\uparrow X}}
\quad 
  m ^{\uparrow X}(A) = 0$$
otherwise. 
The functions  $Bel ^{\uparrow X}$,  $Pl ^{\uparrow X}$  and  $Q ^{\uparrow 
X}$  are defined accordingly to Bel, Pl and Q definitions above with respect 
to $m ^{\uparrow X}$ as their (pseudo-)mass function.\\
\end{df}

Please notice that the operator $\oplus$ is defined for combination of Bel's 
only for the same space of discernment. Should it happen, however that $Bel_1$ 
is defined over the space $X_1 \times X_3$, and $Bel_2$ over 
$X_2 \times X_3$, 
then instead of writing: \\
$$Bel_{1,2}=Bel_1 ^{\uparrow X_1 \times X_2 \times X_3} \oplus 
 B_2 ^{\uparrow X_1 \times X_2 \times X_3} $$
we will simply write
$$Bel_{1,2}=Bel_1  \oplus Bel_1 $$
whenever no misunderstandings may occur.

\section{Prerequisites for Shenoy-Shafer Propagation}

We cite below extensively the paper \cite{Shenoy:90} of Shenoy and Shafer
to recall some basic notions and to show usefulness of             
decomposition of the DS joint belief distribution in terms of a belief 
network..

Variables and valuations:Let V be a finite set. The elements of V are called 
variables. For each $h  \subseteq V$ there is a set $VV_h$. The elements of  
$VV_h$ are  called valuations. Let VV=$\bigcup \{ VV_h | h \subseteq V \}$ be 
called the set of all valuations.\\

 In case of probabilities a valuation on h will be a non-negative, 
real-valued 
function on the set of all configurations  of h(a configuration on h is a 
vector of possible values of variables in h). In the belief function case  a 
valuation is a non-negative, real-valued function on the set of all     
subsets of 
configurations of h.\\

Proper valuation: for each $h \subseteq V$ there is a subset $P_h$ of $VV_h$ 
elements of which are called proper valuations on h. Let P be the set of all 
proper valuations. \\

Combination: We assume that there is a mapping $\odot: VV \times VV 
\rightarrow VV$ called combination such that:\\
(i) if G and H are valuations on g and h respectively, then $G \odot H$ is a 
valuation on $g \cup h$ \\
(ii) if either G or H is not a proper valuation then  $G \odot H$ is not a 
proper valuation \\
(iii) if both G and H are proper valuations then  $G \odot H$ may be or not 
be a proper valuation \\

In case of probabilities, combination is value-by-value multiplication. In 
case of DS-theory - it is the Dempster rule operator $\oplus$ (previous 
section).\\

Marginalization: We assume that there is a mapping $\downarrow h:
\bigcup \{ VV_g| g \subseteq h\} \rightarrow VV_h$ called 
marginalization to h such that:\\
(i) if G is a valuation on g and $h \subseteq g$ then $G ^{\downarrow h}$ 
is a valuation on h. \\
(ii) if G is a proper valuation then   $G ^{\downarrow h}$  is a proper 
valuation \\
(iii) if G is a not proper valuation then   $G ^{\downarrow h}$  is not a 
proper valuation \\

In case of probabilities, marginalization is the summation over 
the dropped  dimension(s). 
In case of DS-theory - it is the Dempster-Shafer marginalization.\\

Axiom A1: (Cummutativity and associativity of combination). Suppose G,H,K are 
valuations on g, h, k respectively. Then $G \odot H=H \odot G$ and $(G\odot 
H)\odot K=G \odot (H \odot K)$.\\
Axiom A2: (Consonance of marginalization) Suppose G is a valuation on g, and 
suppose $k \subseteq h \subseteq g$. Then $(G ^{\downarrow h}) ^{\downarrow 
k}= G  ^{\downarrow k}$\\
Axiom A3: (Distributivity  of marginalization over combination) Suppose G and 
H are valuations on g and h, respectively. Then 
 $(G \odot H) ^{\downarrow g}=G \odot (H  ^{\downarrow g \cap h})$ \\

Hypergraph: We call a non-empty set HV of non-empty subsets 
of a finite set of 
 V a hypergraph on V. We call the elements of HV hyperedges. We call the 
elements of V nodes.

Factorization: Suppose A is a valuation on a finite set of variables V, and 
suppose HV is a hypergraph on V. If A is equal to the combination of 
valuations of all hyperedges h  of HV then we say that A factorizes on HV.
\\
The axiom A3 states that to compute $(G \odot H) ^{\downarrow g}$ it is not 
necessary to compute $G \odot H$ first.\\

Shenoy and Shafer consider it unimportant whether or not the factorization 
should refer to conditional probabilities in case of probabilistic belief 
networks. We shall make at this point the remark that for expert system 
inference engine it is of primary iomportance how the contents of the 
knowledge base should be understood by the  user as beside computation an 
expert system is expected at least to justify its conclusions 
and it can do so only referring to elements of the knowledge base. So if a 
belief network (or a 
hypergraph)
 is to be used as the knowledge base, as much elements as possible 
have to refer to experience of the user.\\

In our opinion, the major reason for this remark of Shenoy and Shafer is that 
in fact the Dempster-Shafer belief function cannot be decomposed in terms of 
conditional belief functions as they are defined in the literature.        
This claim 
is demonstrated in the next section. Thereafter we introduce our own 
definition of conditionality for the DS theory.

\section{Definitions of Conditionality in Literature}

 The probability update function  $cond_B$ with respect to the event 
(evidence) 
B is defined (e.g. in \cite{Halpern:92}) 
as a partial function from the set of
probability
 functions into the 
set of probability functions. As usual, let  $cond_B (Pr) = Pr(.|B)$ 
($Pr$ stands for "probability"). It is 
known that then if   $\circ$ denotes the operator of update combination tthen 
the following holds:
$$cond_C \circ cond_B = cond_{B \cap C}=cond_B \circ cond_C$$

The belief update function be defined
 (after e.g. \cite{Halpern:92}) 
 with respect to evidence B ($cond_B$) 
as a partial function from the set of belief functions into the set of
belief functions.

\subsection{Dempsterian Interpretation of Conditional Belief}

Dempster  \cite{Dempster:67}
defined conditional belief function for a Bel function conditioned  
on the event B as:
 $$Bel(.||B)=Bel \oplus Bel_B$$
(notation after \cite{Halpern:92})
with  $Bel_B$ being the determinictic function of belief into validity 
of the event B, that is $m_B(B)=1$ and $m_B(A)=0$ for every  $A 
\neq B$. It has been shown that 
$$Bel(A||B)= \frac{Bel(A \cup B ^c)- Bel(B ^c)}
                  {1-Bel(B ^c)}$$
which implies directly:
$$Pl(A||B)= \frac{Pl(A \cap B)} {Pl(B)}$$

It is easy to show that for  $cond_B=Bel(.||B)$ the following holds:
 
 $$cond_C \circ cond_B = cond_{B \cap C}=cond_B \circ cond_C$$

\subsection{Halpernian Interpretation of Conditional Belief}

Halpern and  Fagin \cite{Halpern:92}
insisted on treating the belief function as generalized probability. 

Let   P be a set of probability functions defined over a sample space 
  $\Xi$. A lower envelop of P is defined as a function f such that 
for every $A \subseteq \Xi$ \quad f(A)=inf(Pr(A);Pr $\subseteq$ P (Pr means 
probability), 
 A is measurable with respect to   Pr). Thee upper envelop is defined 
respectively
(as   supremum). Let  Bel be a belief function defined over       
  $\Xi$,and let   ( $\Xi$, $\Alpha$, Pr) be a probability space.
We say that Pr is consistent with  Bel, if $Bel(A) \leq Pr(A) 
\leq 
Pl(A)$ for every $A \in \Alpha$. 
This reflects the intuition that Pr is consistent with Bel, 
if the probabilities assigned by Pr are consistent with intervals 
[Bel(A),Pl(A)] set by Bel. For consistency it suffuces that  $Bel(A) \leq 
Pr(A)$, as then immediately $ Pr(A) \leq Pl(A)$. Hence Bel is the lower 
envelop for P, and Pl the upper envelop for P. 

Though every belief function is a lower envelop, not every lower envelop is a 
belief function. 

Let for the belief function  Bel  $P_{Bel}$ denote the set of all probability 
functions consistent with       Bel. 
It has been shown   \cite{Halpern:92}
that if Bel is a belief function over     $\Xi$, then for 
every  A, $A \subseteq \Xi$ we have:
$$Bel(A)=inf_{Pr \in  P_{Bel}} Pr(A)$$
$$ Pl(A)=sup_{Pr \in  P_{Bel}} Pr(A)$$

Halpern and Fagin define conditional belief function with respect to the
event B
$Bel(A|B)$ with respect to the belief functiona Bel()  as:\\
$$Bel(A|B)=inf_{Pr \in  P_{Bel}} Pr(A|B)$$
$$ Pl(A|B)=sup_{Pr \in  P_{Bel}} Pr(A|B)$$

It  has been shown \cite{Halpern:92} that if  Bel() is a belief function such 
that $Bel(B)>0$, then 
$$Bel(A|B)= \frac{Bel(A \cap B)} {Bel(A \cap B)+ Pl(A ^c \cap B)}$$
$$ PL(A|B)= \frac{ Pl(A \cap B)} { Pl(A \cap B)+Bel(A ^c \cap B)}$$
It has been shown further  \cite{Halpern:92} that that then 
 $Bel(.|B)$ is a belief function and $Pl(.|B)$ 
the respective plausibility function.  

One may be tempted by analogy to probability update function
to define  $cond_B=Bel(.|B)$.
However, it has been shown that in general case for the belief functions:
 $$cond_C \circ cond_B \neq cond_{B \cap C} \neq cond_B \circ cond_C$$

\subsection{Kyburgian Definitions of Conditionality}

Kyburg \cite{Kyburg:87} was probably the first to demonstrate 
that a belief function may be represented as an envelop of a family
of traditional probability functions.. He proved also that the following 
holds:
$$Bel(A|B) \leq Bel(A||B) \leq Pl(A||B) \leq Pl(A|B)$$

He considered also non-deterministic evidence. He defined non-deterministic 
conditional probability with respect to event B with probability p as 
(Jeffry-rule):
$$ P(X|^p B)= P(X|B) \cdot p + P(X|B ^c)\cdot (1-p)$$

\noindent (Notice that: $P(B|B)=1$, and $P(B|^p B)=p$),\\
In analogy to the above definitions, he
 defined also two new types of non-deterministic conditional belief 
functions: First  of them in analogy to Dempsterian conditional belief: Let 
$Bel ^{p,B}$, 
be the so-called simple belief function such that $m  
^{p,B}(B)=p$ and  $m ^{p,B}(\Xi)=1-p$, 
and for other subsets of  $\Xi$ m is equal 0, with  $\Xi$ 
being the universe (the space of discernment). Then conditional belief  
$Bel(.||^p B)$ is defined as
 $$Bel(.||^p B)=Bel \oplus Bel ^{p,B}$$ 
that is as Dempsterian combination of belief function and the simple belief 
function.\\
On the other hand, he introduced also a generalized  "envelop" definition of
the conditional belief function:%
$$Bel(A|^p B)=inf_{Pr \in  P_{Bel}} Pr(A|^p B)$$
and subsequently he has shown that \\

$$Bel(A|^p B) \leq Bel(A||^p B) \leq Pl(A||^p B) \leq Pl(A|^p B)$$

\subsection{Criticism of the Notions of Conditionality}

The "envelop" definitions of conditionality ($Bel(.|B)$, $Bel(.|^p B)$ above) 
share one property making then not 
suitable for Shenoy-Shafer propagation: the belief update function is in 
general not cumutative, hence the sequence of usage of evidence proves to be 
of importance, and hence the Axiom A1 is violated. 

It should also be mentioned that Smets \cite{Smets:92} sharply criticizes 
envelop interpretations proposed by Kyburg, Halpern and Fagin as well as 
earlier by Dempster (see above) as misleading and not compatible with the 
spirit of the DST.\\

On the other hand, we would expect of a conditional belief function that :\\
$Bel = Bel (.|| B) \oplus Bel_B$, but obviously this is not the case (in 
general it is not true that: $Bel = Bel  \oplus Bel_B \oplus Bel_B$). \\
Hence it is impossible that the factorization using $Bel(.||B)$ may reflect 
the function $Bel$. \\

Under these circumstances one should plainly ask: why not to try the 
expression
\begin{df} If Bel is a DS belief function defined over some space of
discernment, and h is a set of variables spanning this space, 
then the conditional belief function $Bel  ^{|h}$ should be any function
meeting the condition
 $Bel = Bel ^{|h} \oplus Bel ^{\downarrow h}$ 
\end{df}
as the defining 
expression for conditioning on the set of variables h.\\
There are several severe reasons why this amazingly simple idea may have
not been exploited in the past:\\
(i) in general,  $Bel ^{|h}$ is not unique.\\
(ii)  in general,  $Bel ^{|h}$ is not a belief function (it is only a 
pseudo-belief function, with Q-values being non-negative). \\

      In order to exploit still the results of Shenoy-Shafer, it must be 
stated that 
neither the dempsterian  combination nor dempsterian marginalization leads 
outside the set of pseudo-belief function set. Hence it may be easily shown 
that Shenoy-Shafer axioms apply also to pseudo-belief functions. Given this 
prerequisite, we develop in the subsequent sections a theory of 
identification of DS belief networks following results obtained 
by  Spirtes, Glymour and Scheines \cite{Spirtes:90b} 
for 
probabilistic belief networks. 
\section{An Alternative Definition of Conditionality and Implications}

In \cite{Klopotek:93b} we have introduced a new
frequentist
 interpretation of DS Belief 
function. Under this interpretation we introduced the notion of composite 
measurement of two variables $X_1,X_2$, a notion of statistical independence 
of Bel functions closely related to the empty extension and we introduced a 
new notion of conditionality. Below we briefly summarize this interpretation:

F. Bacchus in his paper \cite{Bacchus:90} on axiomatization of 
probability theory and 
first order logic shows that probability should be considered as a quantifier 
binding free variables
in first order logic expressions just like universal and existential 
quantifiers do. So if e.g. $\alpha(x)$ is an open expression with a free 
variable $x$ then   $[\alpha(x)]_x$ means the probability of truth of the 
expression  $\alpha(x)$. 
(The quantifier $[]_x$ binds the free variable $x$ and yields a numerical 
value ranging from 0 to 1 and meeting all the Kolmogoroff axioms). 
Within the expression  $[\alpha(x)]_x$ the variable 
$x$ is bound. While sharing Bacchus' view, we find his notation a 
bit cumbersome so we change it to be similar to the universal and 
existential quantifiers throughout this paper.
Furthermore, Morgan \cite{Morgan:91} insisted that the probabilities be 
always considered in close connection with the population they refer to.  
 Bacchus' expression 
$[\alpha(x)]_x$ we rewrite as:\\
  $\Prob{x}{P}\alpha(x)$ - the probability of  $\alpha(x)]$ being true within 
the population P. The  P (population) is a unary predicate with P(x)=TRUE 
indicating that the object x($\in \Omega$, that is element of a universe of 
objects) belongs to the population under considerations. If P and P' are 
populations such that $\forall_x P'(x)\rightarrow P(x)$ (that is membership 
in P' implies membership in P, or in other words: P' is a subpopulation of 
P), then we distinguish two cases:\\
case 1: $(\Prob{x}{P}P'(x))=0$ (that is probability of membership in P' with 
respect to P is equal 0) - then (according to \cite{Morgan:91} for any 
expression $\alpha(x)$ in free variable x the following holds for the 
population P': $(\Prob{x}{P'}\alpha(x))=1$\\
case 2: $(\Prob{x}{P}P'(x))>0$then (according to \cite{Morgan:91} for any 
expression $\alpha(x)$ in free variable x the following holds for the 
population P': 
$$(\Prob{x}{P'}\alpha(x))=  \frac {\Prob{x}{P}(\alpha(x) \land P'(x))}
                                {\Prob{x}{P}P'(x)}$$  
We also use the following (now traditional) mathematical symbols:\\
$\forall_{x}\alpha(x)$ - always  $\alpha(x)$ (universal quantifier) \\
$\exists_{x}\alpha(x)$ - there exists an x such that $\alpha(x)$ 
(existential quantifier) \\
\begin{tabular}{lp{7cm}}
$\alpha \land \beta$ & - logical AND of expressions\\
$\bigwedge_{B} \alpha(B)$  & - logical AND over all  instantiations of
the expression $\alpha(B)$ in free 
variable $B$\\
$\alpha \lor \beta$  & - logical OR of expressions\\
$\bigvee_{B} \alpha(B)$  & - logical OR over all  instantiations of
the expression $\alpha(B)$ in free 
variable $B$\\
$\lnot$  & - logical negation\\
$P \cap Q$  & - intersection of two sets\\
$P \cup Q$  & - union of two sets\\
\end{tabular}

\begin{df} \label{MDef}
Let X be a set valued attribute taking its values as subsets of a 
finite domain $\Xi$. By a {\em measurement method} we understand a function:
 $$M: \Omega \times 2^\Xi \rightarrow \{TRUE,FALSE\}$$ 
where $\Omega$ is the set of objects, (or population of objects)
such that 
\begin{itemize}
\item
 $ \forall_{\omega; \omega \in \Omega}
M(\omega,\Xi)=TRUE$ (X takes at least one of values from $\Xi$)
\item
 $ \forall_{\omega; \omega \in \Omega}
M(\omega,\emptyset)=FALSE$ 
\item 
 whenever 
$M(\omega,A)=TRUE$
for $\omega \in \Omega$, $A \subseteq \Xi$
 then for any $B$ such that $A \subset B$ $M(\omega,B)=TRUE$   
holds,
 \item 
 whenever 
$M(\omega,A)=TRUE$
for $\omega \in \Omega$, $A \subseteq \Xi$ and if $card(A)>1$ then there 
exists  $B$, $B \subset A$ such that $M(\omega,B)=TRUE$ holds.
\item 
for every $\omega$ and every $A$
either  
$M(\omega,A)=TRUE$  or 
 $M(\omega,A)=FALSE$ (but never both).
 \end{itemize}
$M(\omega,A)$ indicates that for the object $\omega$ the value of the 
attribute X has a non-empty intersection with the set $A$
 \end{df}

\begin{df}
A {\em label} of an object $\omega \in \Omega$ is a subset of the domain
$\Xi$ of the attribute $X$. \\
A {\em labeling}  under the measurement method $M$  is a function $l: \Omega 
\rightarrow 2^\Xi$ such that for any object  $\omega \in \Omega$ either
$l(\omega)=\emptyset$ or $M(\omega,l(\omega))=TRUE$.\\
Each {\em labelled object}  (under the labeling $l$) 
consists of a 
pair $(O_j,L_j)$, $O_j$ - the j$^{th}$ object, $L_j=l(O_j)$ - its label.\\
By a {\em population  under the labeling $l$} we understand the predicate 
$P:\Omega \rightarrow \{TRUE,FALSE\}$ of the form 
$P(\omega)=TRUE \  iff \ l(\omega) \neq \emptyset$
(or alternatively, the set of objects  for which this predicate is true) \\
 If for every  object of the 
population the label is equal 
 to $\Xi$ then  we  talk  of  an  {\em unlabeled  population} (under the 
labeling $l$), otherwise of a {\em pre-labelled} one.
\end{df}

\begin{df} 
Let $l$ be a labeling under the measurement method $M$. 
Let us consider the population under this labeling.
The modified measurement method 
$$M_l:
 \Omega \times 2^\Xi \rightarrow 
\{TRUE,FALSE\}$$
where $\Omega$ is the set of objects, 
is is defined as  
$$M_l(\omega,A)= M(\omega,A \cap l(\omega) )$$  
(Notice that 
$M_l(\omega,A)=FALSE$ whenever $A \cap l(\omega)= \emptyset$.)
\end{df}

\begin{df}
Let P be a population and $l$ its labeling. Then 

$$Bel_P    ^{M_l}(A)=\Prob{\omega}{P} \lnot M_l(\omega,\Xi-A)$$

$$Pl_P ^{M_l}(A)=\Prob{\omega}{P} M_l(\omega,A)$$

$$m_P ^{M_l}(A)=\Prob{\omega}{P} (\bigwedge_{B;B=\{v_i\}\subseteq A}
 M_l(\omega,B)
  \land \bigwedge_{B;B=\{v_i\}\subseteq \Xi-A} \lnot
 M_l(\omega,B))$$
$$Q_P ^{M_l}(A)=\Prob{\omega}{P} (\bigwedge_{B;B \neq \emptyset, B \subseteq 
A} M_l(\omega,B))$$
\end{df}

\begin{th}
$m_P    ^{M_l}$, $Bel_P    ^{M_l}$, $Pl_P    ^{M_l}$, $Q_P    ^{M_l}$ are
Mass, Belief, Plausibility and Q Functions in the sense of the Dempster-Shafer 
Theory resp. 
\end{th}

\begin{df}
Let $M$ be a measurement method, $l$ be a labeling under this measurement
method, and P be a population under this labeling (Note that the population
may also be unlabeled).
 let  us  take  a  set  of (not  necessarily  disjoint) nonempty sets  of  
attribute values $\{L ^1, L ^2, ...,L ^k\}$    and 
let us define the  probability of selection as a function
$m ^{LP, L ^1, L ^2, ...,L ^k}: 2 ^\Xi \rightarrow [0,1]$ such that
$$\sum_{A;A \subseteq \Xi}m ^{LP, L ^1, L ^2, ...,L ^k}(A)=1$$
$$\forall_{A; A \in \{ L ^1, L ^2, ...,L ^k\}} 
m ^{LP, L ^1, L ^2, ...,L ^k}(A)>0$$
$$\forall_{A; A \not\in \{ L ^1, L ^2, ...,L ^k\}} 
m ^{LP, L ^1, L ^2, ...,L ^k}(A)=0$$
 The  {\em (general) labelling  process}   on    
the
population P 
is defined as a (randomized) functional 
$LP: 2^{2^\Xi} \times \Delta
\times  \Gamma \rightarrow \Gamma$, where $\Gamma$ is the set 
of all  possible labelings under $M$, and $\Delta$ is 
a set of all possible probability of selection functions,
such that for the given labeling $l$ and a given 
 set  of (not  necessarily  disjoint) nonempty sets  of  
attribute values $\{L ^1, L ^2, ...,L ^k\}$    and 
a given probability of selection 
$m ^{LP, L ^1, L ^2, ...,L ^k}$
it delivers a new labeling $l"$ such that for every object
$\omega \in \Omega$:

1. a label L, element of the set $\{ L ^1, L ^2, ...,L ^k\}$ 
is sampled randomly according to the probability distribution 
$m ^{LP, L ^1, L ^2, ...,L ^k}$;
This sampling is done independently for each individual object,

2. if  $M_l(\omega,L)=FALSE$ then  
$l"(\omega)=\emptyset$\\
(that is l" discards an object $(\omega,l(\omega))$ if 
$M_l(\omega,L )=FALSE$ 

3. otherwise $l"(\omega)=l(\omega) \cap L $
(that is l" labels the object with $l(\omega) \cap L $ otherwise.)
\end{df}

\begin{th} 
$m ^{LP,L ^1,...,L ^k}$ is a Mass Function in sense of DS-Theory.
\end{th}

Let   $Bel   ^{LP;L ^1,...,L ^k}$   be   the   belief    and    $Pl 
^{LP,L ^1,...,L ^k}$  be  the 
Plausibility corresponding to $m ^{LP,L ^1,...,L ^k}$. Now let  us  pose  the 
question: what is the relationship between $Bel_{P"} ^{M_{l"}}$, 
 $Bel_P ^{M_l}$,  and $Bel ^{LP,L ^1,...,L ^k}$. It is easy to show that 

\begin{th} 
Let $M$ be a measurement function, $l$ a labeling, P a population under
this labeling. 
Let $LP$ be a generalized labeling process and let $l"$
be the result of application of the $LP$ for the set
of labels from the set $\{ L ^1, L ^2, ...,L ^k\}$ 
 sampled randomly according to the probability distribution 
$m ^{LP, L ^1, L ^2, ...,L ^k}$;.
Let P" be a population under the labeling $l"$.
Then 
the expected value 
over the set of all possible resultant labelings $l"$ (and hence
populations P") 
(or, more precisely, value vector) of 
$Bel_{P"} ^{M_{l"}}$ is a  combination  via  DS  Combination 
rule of  $Bel_P ^{M_l}$,  and $Bel ^{LP,L ^1,...,L ^k}$., that is:
$$E(Bel_{P"} ^{M_l'}) = Bel_P ^{M_l} \oplus Bel ^{LP,L ^1,...,L ^k}$$.
\end{th}

Let us assume that our space  of  discernment consists  of  the 
attributes $X_1,X_2$ ($ X= X_1 \times X_2)$ ranging over 
$\Xi_1=\{v_{11},v_{12},...,v_{1n_1}\}$,
$\Xi_2=\{v_{21},v_{21},...,v_{2n_2}\}$,

Let us understand the marginal distribution of $X_i$ as  follows: 
the measurement method for the subspace $X_i$  be  equal  to  the 
logical sum of all the measurements on  sets  from  X  compatible 
with the given set in $X_i$.
$$M_l ^{\downarrow X_i}(object,A)=
\bigvee_{B;A = B ^{\downarrow X_i} } M_l(object,B) $$
This implies immediately that
$$m_P ^{M_l ^{\downarrow X_i}}(A) = (m_P ^{M_l}) ^{\downarrow X_i}(A)$$
for $A \subseteq \Xi_i$

As the above relationship holds and we are subsequently concerned with
only one population P, we will drop indices referring to the measurement 
method and the population relying only on projections.

Let us now introduce the notion of quantitative independence for DS-Theory. 
\begin{df} \label{DSindepdf}
Two variables $X_1,X_2$  are (mutually, marginally) 
quantitatively
independent when for 
objects of the population
knowledge of the truth value of $M_l ^{\downarrow X_1}
(object,A ^{\downarrow X_1})$ for all 
$A \subseteq \Xi_1 \times \Xi_2$ does not change our prediction capability
of the values  of $M_l ^{\downarrow X_2}
(object,B ^{\downarrow X_2})$ for any
$B \subseteq \Xi_1 \times \Xi_2$, that is
$$\Prob{\omega}{P} M_l ^{\downarrow X_2}(\omega,B ^{\downarrow X_2})=
\Prob{\omega}{ M_l ^{\downarrow X_1}(\omega,A ^{\downarrow X_1}) \land  P} 
M_l ^{\downarrow X_2}(\omega,B ^{\downarrow X_2})
 $$
 \end{df}

\begin{th}
If variables $X_1,X_2$ are quantitatively independent, then 
for any $B \subseteq \Xi_2$, $A \subseteq \Xi_1$
$$
 m  ^{\downarrow X_2}(B) \cdot m ^{\downarrow X_1}(A) =
\sum_{F; F ^{\downarrow X_1}=A,F ^{\downarrow X_2}=B} m(F) $$
\end{th}

\begin{df} Two variables $X_1,X_2$ are measured compositely iff 
for $A \subseteq \Xi_1, B \subseteq \Xi_2$ for every object $\omega$:
$$M(\omega,A \times C) = M(\omega, A \times \Xi_2) \land
M(\omega, \Xi_1 \times C) $$
and whenever $M(\omega, B)$ is sought,
$$M(\omega, B) = \bigvee_{A,C; A \subseteq \Xi_1, C \subseteq \Xi_2,
A\times C \subseteq B}
 M(\omega,A \times B)$$
\end{df}

Under these circumstances, it is easily shown that 
 whenever $m(B) > 0 $, then there  exist 
A and C such that: $B = A \times C$.

So we obtain:

\begin{th}
If variables $X_1,X_2$ are quantitatively independent and measured
compositely, then 
$$m(A \times C)= m ^{\downarrow X_1}(A) \cdot  m ^{\downarrow X_2}(C) 
$$
\end{th}

Hence the Belief function can be calculated from Belief functions 
of independent variables under composite measurement:

\begin{th}
If variables $X_1,X_2$ are quantitatively independent and measured
compositely, then 
$$Bel=Bel ^{\downarrow X_1} \oplus  Bel ^{\downarrow X_2}
$$
\end{th}

Let us justify now the notion of empty extension:

\begin{df}  \label{DSvardistindep}
The joint distribution over $X=X_1 \times X_2$
in variables $X_1,X_2$  is quantitatively
independent of the variable $X_1$  when for objects of the population
for every A,$A \subseteq \Xi_1 \times \Xi_2$
knowledge of the truth value of $M_l ^{\downarrow X_1}
(object,A ^{\downarrow X_1})$ 
 does not change our prediction capability
of the values  of $M_l(object,A )$, that is
$$\Prob{\omega}{P} M_l (\omega,A )=
\Prob{\omega}{ M_l ^{\downarrow X_1}(\omega,A ^{\downarrow X_1}) \land P}
 M_l (\omega,A) 
 $$
 \end{df}
\begin{th}
The joint distribution over $X=X_1 \times X_2$
in variables   $X_1,X_2$,  measured
compositely,
 is 
independent of the variable $X_1$  only if 
$m ^{\downarrow X_2}(\Xi_2)=1$
that is the whole mass of the marginalized distribution onto $X_2$ is 
concentrated at the only focal point 
$\Xi_2$. 
\end{th}

\begin{th}
If for $X=X_1 \times X_2$ 
$Bel=(Bel ^{\downarrow X_2})^{\uparrow X}$
that is 
Bel is the empty extension of some Bel defined  only over $X_2$,
then the Bel is independent of the variable $X_2$.\\
If for a Bel over  $X=X_1 \times X_2$ with $X_1,X_2$ measured compisitely
Bel is independent of $X_2$, then $Bel=(Bel ^{\downarrow X_2})^{\uparrow X}$.
\end{th}

In the light of the above theorem, and taking into account that a belief 
function which is an empty extension of another function  may be stored in a 
compressed manner, we shall say 

\begin{df}
Let Bel be defined over $X_1 \times X_2$.
We shall speak that Bel is {\em compressibly independent} of $X_2$ iff 
$Bel=(Bel ^{\downarrow X_1}) ^{\uparrow X_2}$. 
\end{df}

REMARK: $m  ^{\downarrow X_1}(\Xi_1)=1$ does not imply empty extension as 
such, especially for non-sigleton values of the variable $X_2$. As previously
with marginal independence, it is the composite measurement that 
makes the empty extension a practical notion.\\

Let us consider now the conditional independence:

Let us introduce a concept of conditionality related  to  the 
above definition of independence.  Traditionally,  conditionality 
is introduced to obtain a kind of independence between  variables 
de facto on one another. So let us define that:

\begin{df}
For discourse spaces of the form $X=X_1 \times ... \times X_n$
we define conditional belief function 
$Bel ^{X | X_i}(A)$ as 
 $$Bel=Bel ^{\downarrow X_i} \oplus  Bel ^{X | X_i}$$
\end{df}

Let us notice at this point that the conditional belief as defined above does 
not need to be unique, hence we have here a kind of pseudoinversion
of the $\oplus$ operator.  Furthermore, the conditional belief does not need 
to 
be a belief function at all, because some focal points m may be negative. 
But it is then the pseudo-belief function in the sense of the DS-theory as 
the Q-measure remains positive. 
Please recall the fact that if $Bel_{12}=Bel_1 \oplus Bel_2$ then
$Q_{12}(A)=c\cdot Q_1(A)\cdot Q_2(A)$, c being a proportionality factor
(as all supersets of a set  are contained in all intersections of its 
supersets 
and vice versa). Hence also for 
our conditional belief definition: 
 $$Q(A)=c \cdot  (Q ^{\downarrow X_i})  ^{\uparrow X}(A) \cdot  Q ^{X | 
X_i}(A)$$ 
We shall talk later of unnormalized conditional belief $Q_*^{X | X_i}$ iff\\
 $$ Q_*  ^{X | X_i}(A) = Q(A)/(Q ^{\downarrow X_i})  ^{\uparrow X}(A) $$ 
 Let us now reconsider the problem of independence, this time of a 
conditional distribution of $(X_1 \times X_2 \times X_3 | X_1 \times X_3)$ 
from the third variable $X_3$.

\begin{th}
Let $X =X_1 \times X_2 \times X_3 $ and let $Bel$ be defined 
 over X.
Furthermore let $Bel ^{X|X_1 \times X_3}$ be a conditional Belief conditioned 
on variables $X_1,X_3$. Let this conditional distribution be
 compressibly 
 independent of 
$X_3$. Let $Bel ^{\downarrow X_1 \times X_2}$ be the projection of $Bel$ onto 
the subspace spanned by $X_1,X_2$. Then there exists 
 $Bel ^{\downarrow X_1 \times X_2 | X_1}$  being a conditional belief of that 
projected belief conditioned on the variable $X_1$ such that this  $Bel 
^{X|X_1 \times 
X_3}$  is the empty extension of  $Bel ^{\downarrow X_1 \times X_2 | X_1}$ 
 $$Bel ^{X|X_1 \times X_3} =
  (Bel ^{\downarrow X_1 \times X_2 | X_1}) ^{\uparrow X}$$
\end{th}

Let us notice that under the conditions of the above theorem

 $$ Bel = 
Bel ^{X|X_1 \times X_3}\oplus Bel ^{\downarrow X_1 \times X_3 } =
  Bel ^{\downarrow X_1 \times X_2 | X_1} 
\oplus Bel ^{\downarrow X_1 \times X_3 }
$$

and hence for any $Bel ^{\downarrow X_1 \times X_3 | X_1}$
$$Bel =  
  Bel ^{\downarrow X_1 \times X_2 | X_1}
\oplus Bel ^{\downarrow X_1}
\oplus Bel ^{\downarrow X_1 \times X_3 | X_1 }
$$

and therefore
$$Bel =  
  Bel ^{\downarrow X_1 \times X_2} 
\oplus Bel ^{\downarrow X_1 \times X_3 | X_1 }
$$
This means that whenever the conditional 
$Bel ^{X_1 \times X_2 \times X_3|X_1 \times X_3}$ 
is compressibly independent of $X_3$, 
then there exists a 
conditional 
$Bel ^{X_1 \times X_2 \times X_3|X_1 \times X_2}$ 
compressibly independent of $X_2$.
But this fact  combined with the previous theorem results in:

\begin{th}
Let $X =X_1 \times X_2 \times X_3$ and let $Bel$ be defined 
over X.
Furthermore let $Bel ^{X|X_1 \times X_3}$ be a conditional Belief conditioned 
on variables $X_1,X_3$. Let this conditional distribution be 
compressibly independent of 
$X_3$. 
Then the empty extension onto $X$ of any 
 $Bel ^{\downarrow X_1 \times X_2 | X_1}$  being a conditional belief of 
projected belief conditioned on the variable $X_1$ 
is a conditional belief function of $X$  conditioned 
on variables $X_1,X_3$. Hence for every $A\subseteq \Xi$
$$ \frac{ Q(A) }
{ Q ^{\downarrow X_1 \times X_3}(A ^{\downarrow X_1 \times X_3} ) }
=  \frac { Q ^{\downarrow X_1 \times X_2}(A ^{\downarrow X_1 \times X_2} ) }
{ Q ^{\downarrow X_1}(A ^{\downarrow X_1} ) }
$$
\end{th}


In this way we obtained some sense of conditionality suitable for 
decomposition of a joint belief distribution.

Above we defined precisely what is meant by marginal 
independence of two variables in terms of the relationship between marginals 
and the joint distribution, as well as concerning the independence of a joint 
distribution from a single variable.\\

For the former case we can establish frequency tables  with  rows 
and columns 
corresponding to cardinalities of focal points of the first and the second 
marginal, and inner elements being cardinalities from the respective sum on 
DS-masses of 
the joint distribution. Clearly, cases falling into different inner 
categories of the table are different and hence $\chi ^2$ test is 
applicable.\\
The match can be $\chi ^2$-tested. The following 
formula should be followed for calculation 
$$\sum_{A;A \subseteq \Xi_1, m ^{\downarrow X_1}(A)>0}
  \sum_{B;B \subseteq \Xi_2, m ^{\downarrow X_2}(B)>0}
\frac { ((
\sum_{C;C \subseteq \Xi, A = C ^{\downarrow X_1}
B=C  ^{\downarrow X_2}}
m(C))
- m ^{\downarrow X_1}(A)\cdot  m ^{\downarrow X_2}(B))^2
}
{ m ^{\downarrow X_1}(A)\cdot  m ^{\downarrow X_2}(B)}$$

The number of df is calculated as
$$ (card(\{A;A \subseteq \Xi_1, m ^{\downarrow X_1}(A)>0\}) -1)\cdot 
(card(\{B;B \subseteq \Xi_2, m ^{\downarrow X_2}(B)>0\})-1)$$


In case of independence of a distribution from one variable one needs to 
calculate the marginal of the distribution of that variable, say $X_i$.
Then the measure of discrepancy from the assumption of independence is given 
as:
$$ 1 - m ^{\downarrow X_i}(\Xi_i)$$
Statistically we can test, based on Bernoullie distribution, what is the 
lowest possible and the highest possible value of $ 1 - m ^{\downarrow 
X_i}(\Xi_i)$ for a given significance level of the true underlying 
distribution.\\

In  case of 
independence between the conditional distribution and one of conditioning
variables, however, it is useless to calculate the pseudoinversion of 
$\oplus$, as we are working then with a population and a sample the size of 
which is not properly defined (by the "anti-labeling").
  But we can build the contingency table of the unconditional joint 
distribution for the independent variable on 
the one hand  and the remaining variables on the other hand, and compare the
respective cells on how do they match the distribution we would obtain 
assuming the  independence. So let m be a Mass Function for the variable 
$X=X_1 \times X_2 \times X_3$. Composite measurement of $X_1,X_2, X_3$ is to 
be assumed. We want to show that $X_1$ conditioned on $X_2$ is independent of 
$_3$. We calculate Q of $m$, and $Q ^{\downarrow X_2}$,       $Q ^{\downarrow 
X_1 \times X_2}$,       $Q ^{\downarrow X_2 \times X_3}$.
We define $Q_t$ to be a Q-function calculated as follows:
$$Q_t=c \cdot Q ^{\downarrow X_2 \times X_3} \cdot
\frac { Q ^{\downarrow X_1 \times X_2}}
{Q ^{\downarrow X_2}}
$$
c - a normalizing constant. Let $m_t$ be the (Pseudo-)Mass Function 
corresponding to $Q_t$. Should $m_t(A)$ have a negative mass for any set $A 
\subseteq \Xi$, so the hypothesis of independence should be viewed as 
statistically rejected. Also if  $m_t(A)=0$ and  $m(A)>0$ holds for any  $A 
\subseteq \Xi$, then   hypothesis of independence should be viewed as 
statistically rejected.
Otherwise 
 we calculate the following $\chi$-statistics: 
$$\sum_{A;A \subseteq \Xi m_t(A)\neq0}
\frac { (m(A)-m_t(A))^2}{|m_t(A)|}$$

The number of degrees of freedom for the      
$\chi ^2$ 
test would then be 
the product: 
(the number of focal points of the projection of the joint distribution
onto    $X_1 \times X_2$  minus one) *  
(the number of focal points of the projection of the joint distribution
onto    $X_2 \times X_3$  minus one).  
%
%

 
\section{DS Belief Network - Definition and  Properties}

\begin{df}
 A 
DS Belief 
 network is a pair (D,Bel) where D is a dag (directed acyclic graph)
and Bel  is a DS belief 
distribution called the {\em underlying distribution}. Each node i in D 
corresponds to a variable $X_i$  in Bel, a set of nodes I corresponds to a 
set of variables $X_I$ and $x_i, x_I$
 denote values drawn from the domain of $X_i$ 
 and from the (cross product) domain of $X_I$ respectively. Each node in the 
network  is regarded as a storage cell for any  distribution 
$Bel ^{\downarrow \{X_i\} \cup X_{\pi (i)} |  X_{\pi (i)} }$
 where $X_{\pi (i)}$ is a set of nodes corresponding to 
the 
parent nodes $\pi(i)$ of i.  The underlying distribution represented by a 
DS belief network is computed via:
$$Bel  = \bigoplus_{i=1}^{n}Bel ^{\downarrow \{X_i\} \cup X_{\pi (i)} |  
X_{\pi (i)} } $$
\end{df}

Please pay attention to the expression {\em any distribution} in front of the 
conditional distribution as more than one conditional distribution is 
possible. We may well imagine a situation where the decomposition of a joint 
belief distribution may be valid for some and not for the other set of 
conditional beliefs. Some important properties follow from this definition:\\

\begin{th}
Let DSN=(D,Bel) be a belief network with Bel equal to 
$$Bel  = \bigoplus_{i=1}^{n}Bel ^{\downarrow \{X_i\} \cup X_{\pi (i)} |  
X_{\pi (i)} } $$
Let j be a node in D without any outcoming edges (a terminal node). Then the 
following holds: \\
$$Bel ^{\downarrow \{X_1,...,X_n\} - \{X_j\} }
  = \bigoplus_{i=1,...,n, i \neq j}Bel ^{\downarrow \{X_i\} \cup X_{\pi 
(i)} | X_{\pi (i)} } $$
\end{th}

\AnfBeweis
Let $Bel_2$ be a pseudo-belief function defined over the set of variables $g 
\cup h$, and $Bel_1$ be a pseudo-belief function 
 defined over $g \cup \{ X\}$,  $X \not\in g$, $X \not\in h$, $g \cap h = 
\emptyset$. Let $Bel_{12}=Bel_1 \oplus 
Bel_2$ be a belief function (defined over  $g \cup \{ X\} \cup h$. Let us 
make the projection  $Bel_{12} ^{\downarrow g \cup h}  =(Bel_1 \oplus 
Bel_2) ^{\downarrow g \cup h} $\\
If we investigate the m-values we will find out that:
 $Bel_{12} ^{\downarrow g \cup h}  =Bel_1  ^{\downarrow g}  \oplus 
Bel_2 ^{\downarrow g \cup h} $\\

On the other hand we know that a complete set of Q-values in the DS theory 
determines completely the the Belief Function. \\

Let us assume $Bel_1$  be  a conditional 
belief  $Bel_{12} ^{\downarrow g \cup \{X\} | g}$, and let us assume that the 
equation holds for any such conditional belief. Then 
 $Bel_{12} ^{\downarrow g}  =Bel_1  ^{\downarrow g}  \oplus 
Bel_{12} ^{\downarrow g}$\\
It is easily checked that for non-zero Q-points of $Bel_{12} ^{\downarrow g}$ 
(in this case focal points and 
their subsets) 
$Q_1  ^{\downarrow g}$ must be equal 1. 
This implies, however, that $(Q_1  ^{\downarrow g})^{\uparrow g \cup h}$ must 
be equal 1 also for non-zero Q-points of 
 As any conditional belief is allowed 
then also ones with non-zero Q-points at zero-Q-points of  
$Bel_{12} ^{\downarrow g \cup h}$.\\

But then $Bel_2 ^{\downarrow g \cup h}=Bel_2$ is completely determined to be 
identical with 
$Bel_{12} ^{\downarrow g \cup h}$ both at zero and non-zero Q-points.

What is more, $Bel_2$ is a belief function (and not only a pseudo-belief 
function).
\EndBeweis

This is an important theorem as it states that a properly chosen subnetwork 
reflects a sub Belief function. 

\begin{df}  \cite{Geiger:90} 
A {\em trail } in a dag is a sequence of links that form a path in the 
 underlying undirected graph. A node $\beta$ is called a {\em head-to-head 
node}  with 
respect to a trail t if there are two consecutive links $\alpha \rightarrow 
\beta$ and $\beta \leftarrow \gamma$ on that t. 
\end{df}

\begin{df}  \cite{Geiger:90} 
A trail t connecting nodes $\alpha$ and $\beta$ is said to be 
{\em active } 
given a set of nodes L, if 
(1) every head-to-head-node wrt t either is or has 
a descendent in L and (2) every other node on t is outside L. Otherwise t is 
said to be {\em blocked } (given L).
\end{df}

\begin{df}  \cite{Geiger:90} 
If J,K and L are three disjoint sets of nodes in a dag D, then L is said to 
{\em d-separate } J from K, denoted $I(J,K|L)_D$  iff no trail between a node 
in J and a node in K is active given L.
\end{df}

It has been shown in \cite{Geiger:90b} that 
\begin{th}
Let L be a set of nodes in a dag D, and let $\alpha,\beta \notin L$ be two 
additional nodes in D. Then $\alpha$ and $\beta$ are connected via an active 
trail  (given L) iff  $\alpha$ and $\beta$ are connected via a simple (i.e. 
not possessing cycles in the underlying undirected graph) active trail (given 
L).
\end{th}

We claim that:
\begin{th} 
Let n be a node in a dag D. Let D' be a subgraph of D such that
all (and only) outcoming edges of n are removed. Let L be a set of nodes in 
the dag D (and hence D') containing n ($n \in L$), and let $\alpha,\beta 
\notin L$ be two 
additional nodes in D. Then $\alpha$ and $\beta$ are connected in D via an 
active trail  given L iff  $\alpha$ and $\beta$ are connected via an active 
trail given L in D'.
 \end{th}
\AnfBeweis
As n is in L, the removal of outgoing edges does not influence any active 
trail as n would block any trail containing them.
\EndBeweis

\begin{df}  
If $X_J,X_K,X_L$ are three disjoint sets of variables of a distribution Bel, 
then $X_J,X_K$ are said to be conditionally independent given $X_L$ (denoted 
$I(X_J,X_K |X_L)_{Bel}$ iff 
 $$Bel ^{\downarrow X_J \cup X_K \cup X_L |  X_L} 
 \oplus Bel ^{\downarrow   X_L } =
 Bel ^{\downarrow X_J  \cup X_L |  X_L} \oplus
 Bel ^{\downarrow X_K \cup X_L |  X_L} 
 \oplus Bel ^{\downarrow   X_L } $$
%
$I(X_J,X_K |X_L)_{Bel}$ is called a {\em 
(conditional independence) statement}
\end{df}

\AbbZwoelf

\begin{th} \label{IDIBel}
Let $Bel_D=\{Bel|$(D,Bel) be a DS belief network\}. Then:\\

$I(J,K|L)_D$ iff $I(X_J,X_K |X_L)_{Bel}$ for all $Bel \in Bel_D$.
\end{th}
\AnfBeweis
The "only if" part (soundness) states that whenever  $I(J,K|L)_D$ holds in D, 
it must represent an independence that holds in every underlying
 distribution. We prove it as follows:
Let us take a node l in L having no predecessor in L. Let us try to calculate 
the conditional distribution on l.  if n were a root node (without 
incoming edges) then simply $$Bel ^{|X_l}  = \bigoplus_{i=1,..,n,i\neq l}Bel 
^{\downarrow \{X_i\} \cup X_{\pi (i)} |  X_{\pi (i)} } $$
Otherwise we have to transform the node l into such one. First let us exploit 
the previous theorem and remove all nodes not being predecessors of l and not 
l itself from the graph. The remaining dag represents the joint distribution 
projected onto the remaining nodes. Let us take the youngest predecessor of l 
    (that is a node l' not having a successor which were predecessor of l). 
Let us 
consider the two factors: 
$Bel ^{\downarrow \{X_l\} \cup X_{\pi (l)} |  X_{\pi (l)} } \oplus
Bel 
^{\downarrow \{X_{l'}\} \cup X_{\pi (l')} |  X_{\pi (l')} }$.
It is easy to check that the above DS combination is equal to:
$Bel ^{\downarrow \{X_l,X_l'\} \cup X_{\pi (l)} \cup X_{\pi (l')} |  X_{\pi 
(l)-\{l'\}} \cup  X_{\pi (l')} } $
The above expression may be easily transformed into equivalent 
$Bel ^{\downarrow \{X_l,X_l'\} \cup X_{\pi (l)} \cup X_{\pi (l')} |  X_{\pi 
(l)-\{l'\}} \cup  X_{\pi (l')} }
= Bel ^{\downarrow \{X_l,X_l'\} \cup X_{\pi (l)} \cup X_{\pi (l')} |  X_{\pi 
(l)-\{l'\}} \cup  X_{\pi (l')} \cup \{X_l\} }
\oplus 
Bel ^{\downarrow \{X_l\} \cup X_{\pi (l)} \cup X_{\pi (l')} |  X_{\pi 
(l)-\{l'\}} \cup  X_{\pi (l')} } $
This means that the node l' can be made now a terminal node     and the 
process of 
node removal may be continued until l becomes a root  node (that is a node 
without predecessors). 

Let us consider now the respective graph transformations. The outgoing edges 
of the node l can be removed as shown previously without deactivation of 
active trails and without introducing new ones
 (see e.g. Fig.\ref{abbzwoelf}). 
The change of direction of 
the edge (l',l) with introduction of new edges does not affect active trails 
either (some of them are only shortened, a head-=to-head-meeting is 
by-passed - see e.g. Fig.\ref{abbdreizehn}). Hence we can move all the 
nodes of L to become either 
root        nodes or to have only nodes from L as predecessors. If we remove 
now these nodes from the transformed graph, then the remaining graph will 
represent the conditional distribution on these nodes. And all the active 
trails will not contain any head-to-head meeting. Hence two nodes $\alpha$  
and $\beta$ not connected by an active trail will neither possess 
a common 
predecessor nor be a successor of one another. Let us remove stepwise 
terminal nodes not being $\alpha$,$\beta$. A graph consisting of two disjoint 
graphs with  $\alpha$,$\beta$ as solely terminal nodes of each. Then 
obviously their calculations of marginals may be separated in the remaining 
dag. Hence missing active trail implies independence statement.


\AbbDreizehn

The "if" part (completeness) asserts that any independence that is not 
detected by d-separation cannot be shared by all distributions in $P_D$ and 
hence cannot be revealed by non-numeric methods. 
we prove it by construction of an example  as follows: 
If there exists an active p-trail connecting nodes i,j then 
by graph transformations (edge removal and edge reversal, thereafter terminal 
node removal) as described above 
we obtain a final graph for which either i and j have a common predecessor k 
or there exists an oriented path connecting both i,j. Let $X_i$, $X_j$, $X_k$ 
be variables associated with nodes i,j,k. \\
In the first case (common predecessor) let $Y_1$,...,$Y_m$ be variables 
associated with nodes on the directed path from k to i,     let 
$Z_1$,...,$Z_n$ be 
variables associated with nodes on the directed path from k to j.
Furthermore, let the conditional beliefs associated with the nodes be 
constructed as follows: the only focal points are (c - normalizing 
constants)\\
Node k: \\
$$m ^{\downarrow \{X_k\}\cup X_{\pi(k)}| X_{\pi(k)}}
(X_k=\{v_1\},.....)=p/c_{X_k}$$
$$m ^{\downarrow \{X_k\}\cup X_{\pi(k)}| X_{\pi(k)}}
(X_k=\{v_2\},.....)=(1-p)/c_{X_k}$$
Nodes on the path from k to i (r=1,...,m+1, $Y_0$ means $X_k$, $Y_{m+1}$ 
means $X_i$):\\
$$m ^{\downarrow \{Y_r\}\cup Y_{\pi(r)}| Y_{\pi(r)}}
(Y_r=\{v_1\},Y_{r-1}=\{v_1\},.....)=1/c_{Y_r}$$
$$m ^{\downarrow \{Y_r\}\cup Y_{\pi(r)}| Y_{\pi(r)}}
(Y_r=\{v_2\},Y_{r-1}=\{v_2\},.....)=1/c_{Y_r}$$
Nodes on the path from k to j (r=1,...,n+1, $Z_0$ means $X_k$, $Z_{n+1}$ 
means $X_j$):\\
$$m ^{\downarrow \{Z_r\}\cup Z_{\pi(r)}| Z_{\pi(r)}}
(Z_r=\{v_1\},Z_{r-1}=\{v_1\},.....)=1/c_{Z_r}$$
$$m ^{\downarrow \{Z_r\}\cup Z_{\pi(r)}| Z_{\pi(r)}}
(Z_r=\{v_2\},Z_{r-1}=\{v_2\},.....)=1/c_{Z_r}$$

It is immediately visible that the joint belief distribution of $X_i$ and its 
predecessors in the remaining graph can be expressed as the only focal 
points:\\
$$m ^{\downarrow X_i and its predecessors}(X_i=\{v_1\},X_k=\{v_1\},...)=
p * m ^{\downarrow predecessors of X_i}(X_k=\{v_1\},...)/c$$
$$m ^{\downarrow X_i and its predecessors}(X_i=\{v_2\},X_k=\{v_2\},...)=
(1-p) * m ^{\downarrow predecessors of X_i}(X_k=\{v_2\},...)$$
Hence obviously:\\
$$m ^{\downarrow X_i}(X_i=\{v_1\})=p$$
$$m ^{\downarrow X_i}(X_i=\{v_2\})=(1-p)$$
In the same way we show that:\\
$$m ^{\downarrow X_j}(X_j=\{v_1\})=p$$
$$m ^{\downarrow X_j}(X_j=\{v_2\})=(1-p)$$
and that \\
$$m ^{\downarrow \{X_i,X_j\}}(X_i=\{v_1\},X_j=\{v_1\})=p$$
$$m ^{\downarrow \{X_i,X_j\}}(X_i=\{v_2\},X_j=\{v_2\})=(1-p)$$
But we see immediately, that if\\
$$Bel_{prod}=Bel  ^{\downarrow X_i}\oplus Bel  ^{\downarrow X_j}$$\\
then focal points of $Bel_{prod}$ are\\
$$m_{prod}(X_i=\{v_1\},X_j=\{v_1\})=p ^2$$
$$m_{prod}(X_i=\{v_1\},X_j=\{v_2\})=p*(1-p)$$
$$m_{prod}(X_i=\{v_2\},X_j=\{v_1\})=p*(1-p)$$
$$m_{prod}(X_i=\{v_2\},X_j=\{v_2\})=(1-p) ^2$$
 which is obviously different from $Bel ^{\downarrow \{X_i,X_j\}}$. This 
means, 
however, that for any dependence in the sense of d-separation   we are 
actually capable to construct a joint belief distribution from the family of 
compatible distributions such that there is a dependence in the distribution 
corresponding to the d-separation dependence.\\
In the same manner we can proceed in case of a direct oriented path from i to 
j or from j to i. Again we will manage to construct a belief function where 
missing d-separation at a given point indicates dependence in the 
distribution.
\EndBeweis

\section{Principles for Construction of dag from Data}

Many writers have connected causality with statistical dependence. 
We parallel here  \cite{Spirtes:90b} in formulating the following principles, 
while understanding independence as given by definitions Def.\ref{DSindepdf} 
and \ref{DSvardistindep}

 \begin{df}
Let \V be a set of  random variables with a joint 
DS-belief distribution. We say that variables X,Y $\in$\V are
 {\em directly causally dependent} if and 
 only if there is a causal dependency between X,Y (either the value of X 
influences the value of Y or the value of Y influences the value of X or the 
value of a third variable not in \V influences the values of both X and Y)
that does not involve any other variable in \V.
\end{df}

{\bf Principle I: }  For all X,Y in \V, X and Y are directly causally 
dependent 
if and only if for every subset \SS of \V not containing X or Y, X and Y are 
not statistically independent conditional on \SS.\\

\begin{df}
We  say that {\em B is directly causally dependent on A} provided that A and 
B 
are causally dependent and the direction of causal influence is from A to 
B.\\
\end{df}

{\bf Principle II: } if A and B are directly causally dependent and B and C 
are directly causally dependent, but A and C are not, then:
B is causally dependent on A, and B is causally dependent on C if and only if 
A and C are statistically dependent conditional on any set of variables 
containing B and not containing A or C.\\

{\bf Principle III: } A directed acyclic graph represents a DS-belief
distribution on the variables that are vertices of the graph if and only if\\
for all vertices X,Y and all sets \SS   of vertices in the graph
(X,Y $\notin$ \SS), \SS 
d-separates 
X and Y if and only if X and Y are independent conditional on \SS.\\

Please notice that Principle III bears close resemblance with theorem 
\ref{IDIBel}. It actually transfers a fine property of family of 
belief networks into a criterion for building a belief network. A weak point 
of such a criterion is that a decision whether or not an edge is to be 
included into the underlying dag is based on the whole (at 
the moment of edge inclusion unknown)  structure of the target dag.\\
Fortunately, as we will show below, Principle III implies both 
Principles I and II. This means that Principles I and II, being local with 
respect to the target dag, may provide useful initial hints for construction 
of the target dag. What is more, Principles I and II combined with dag 
definition imply Principle III which means that we can indeed construct the 
whole dag structure exploiting only local properties of the target dag. \\

Let us introduce some notions. First let us define a {\em partially oriented 
graph} (pog)
as a structure (\V,E,O), with \V being the set of nodes, E being the set of 
edges with an edge being a subset of \V with cardinality 2, 
O:E$\rightarrow 2 ^ {V \times V}$ being the 
orientation function of edges assigning each edge $\{X_i,X_j\}$ in E either 
the orientation $\{\}$ (no orientation) or  $\{(X_i,X_j)\}$  (from $X_i$ to 
$X_j$), or   $\{(X_j,X_i)\}$  (from $X_j$ to 
$X_i$) or   $\{(X_i,X_j),(X_j,X_i)\}$  (both from $X_i$ to 
$X_j$ ) and from $X_j$ to  $X_i$).  The  last  orientation  is  an 
unpleasant one, 
but may occur in processes described below, If the first (empty) orientation 
is assigned, the edge is called unoriented, otherwise it is called   
oriented.

Furthermore let us call two edges {\em neighbouring edges} iff they share a 
vertex.   Let $\{X_i,X_j\}$ and  $\{X_k,X_j\}$ be neighbouring edges (they 
share $X_j$ so they are neighbouring at $X_j$). We call them {\em bridged 
edges} iff  there exists an edge  $\{X_i,X_k\}$ in E. Otherwise they are 
called {\em unbridged}. The edge  $\{X_i,X_j\}$  (with respect to the 
neighbouring pair of edges) is said to be {\em head-to-neighbour}
 oriented iff  $(X_i,X_j) \in O( \{X_i,X_j\})$.   The edge    
$\{X_i,X_j\}$ 
(with respect to the 
neighbouring pair of edges) is said to be {\em tail-to-neighbour}
 oriented iff  $(X_j,X_i) \in O( \{X_i,X_j\})$.  

\AbbEins
We claim the following:

\begin{th} \label{iiiDSi} 
Let Bel be a DS-belief   distribution represented by an acyclic directed 
graph 
G according to Principle III. Then G is an orientation (G has the undirected 
structure) of the undirected graph U that represents Bel according to 
Principle I.
\end{th}
\AnfBeweis
 
If two nodes/variables $X_i$ and 
$X_j$ are connected via an undirected edge within the U-graph generated by 
Principle I, then there exists no set of variables $Y_1,....,Y_k$ such that 
for every combination of values 
 $Bel ^{\downarrow \{X_i, X_j, Y_1,...,Y_k\} | \{ Y_1,...,Y_k\}} 
 \oplus Bel ^{\downarrow   \{ Y_1,...,Y_k\} } =
 Bel ^{\downarrow \{X_i, Y_1,...,Y_k\} |\{ Y_1,...,Y_k\}} \oplus
 Bel ^{\downarrow \{X_j, Y_1,...,Y_k\} |\{ Y_1,...,Y_k\}} 
 \oplus Bel ^{\downarrow   \{ Y_1,...,Y_k\}} $
as otherwise the edge would not be 
inserted. Assume for a moment Principle III would not generate a directed 
edge connecting both variables in a directed graph D. Then in this D-graph 
a d-separation of both variables can be found: take simply the set of nodes 
which directly precede any of the variables. But this would enforce 
conditional independence in contradiction with the result established 
previously. So any edge generated by Principle I is also present in every 
graph generated by Principle III. \\

\AbbZwei

On the other hand if Principle I establishes that there is no undirected edge 
connecting both variables then there exists a set of variables on which these 
two are conditionally independent. But then Principle III cannot establish an 
edge between them as there would exist no d-separation between them. So 
whenever Principle I establishes no edge between variables, no edge will be 
established by Principle III.
\EndBeweis
\begin{th}  \label{iiiDSii} 
Principle III implies Principle II.
\end{th}
\AnfBeweis
 Let us consider the graph U generated by Principle I. Let us consider 
partial orientations of the  graph U generated from it by Principle II. It is 
easily seen that there may be only one such orientation. Let us turn our 
attention to  Theorem  \ref{iiiDSii}. 
Let us consider a head-to-head meeting of directed edges $(X_i,X_l)$, 
$(X_j,X_l)$ generated by Principle II, that is $X_i,X_j$ not being 
directly 
connected in U,  $X_i,X_l$ being directly connected in U, 
$X_j,X_l$  being directly connected in U, no set containing $X_l$ rendering 
$X_i,X_j$ independent. Then Principle III has also to generate this 
head-to-head meeting as the existence of the trail of directed edges  
$(X_i,X_l)$, $(X_j,X_l)$ guarantees in this case that no d-separation 
 containing $X_l$ exists.  So every head-to-head-meeting generated by 
Principle 
II occurs also in every graph generated by Principle III. On the other hand, 
if during testing  independence by means of Principle II for the edges  
$(X_i,X_l)$, $(X_j,X_l)$  a set containing $X_l$ was detected such that it  
renders $X_i,X_j$ independent, then head-to-head meeting of these edges 
must not occur if Principle III is applied.
\EndBeweis


In this way we have established that: if there exists a dag of the 
distribution generated by Principle III, then application of Principles I 
and II will deliver its undirected structure and orientation of all  those 
unbridged pairs of 
arcs which meet head-to-head at a node. 
(So if the intrinsic graph is given by Fig.\ref{abbeins} then Principle II 
yields a graph given by Fig.\ref{abbdrei}).

\AbbDrei

Let us now discuss which orientations of other arcs are established rigidly 
by Principle III. Pearl's definition of d-separation refers to arc 
orientation at following nodes:
(1) head-to-head nodes 
(2) direct and indirect descendants of head-to-head nodes 

So let us establish the following principle:\\

{\bf Principle II$^c$} Let H be a partially oriented graph 
generated by Principles I and II. 
Whenever  
  $\{X_i,X_j\}$ and  $\{X_k,X_j\}$ are neighbouring unbridged edges, with
 $\{X_i,X_j\}$ being head-to-neighbour oriented and    $\{X_k,X_j\}$ being 
unoriented, orient    $\{X_k,X_j\}$  tail-to-neighbour. \\

Please notice that Principle  II$^c$ is a kind of operationalization of 
Principle II, as it is a direct consequence of the "if and  only if" 
expression in   Principle II. It has been introduced because the 
formulation of Principle II directs our attention to orienting edges 
head-to-head, but it is less obvious that it also implies some head-to-tail 
orientations.

Obviously, the following theorem holds: 
\begin{th}  \label{iiiDSiic} 
Principle III implies Principle II$^c$.
\end{th}

The Theorem is obvious if we consider the previous ones.
(So if the intrinsic graph is given by Fig.\ref{abbeins} then Principle  
II$^c$ yields a graph given by Fig.\ref{abbvier}).

\AbbVier

Furthermore let us introduce the following principle:

{\bf Principle IV: } Let  H be a partially oriented graph 
generated by Principles I and II  and II$^c$.  Let the subgraph H' 
of H contain only oriented edges in H. Let $\{X_i,X_j\}$  be an unoriented 
edge in H. If $X_j$ is a descendent of $X_i$ in H', then orient this edge 
from $X_i$ to $X_j$. Apply thereafter Principle II$^c$ exhaustively. \\

Obviously
\begin{th}  \label{iiiDSiv} 
Dag-structure and Principle III imply Principle IV.
\end{th}

(So if the intrinsic graph is given by Fig.\ref{abbeins} then Principle IV 
yields a graph given by Fig.\ref{abbfuenf}).\\

\AbbFuenf

{\bf Principle V:} Let  H be a partially oriented graph 
generated by Principles I and II.  
Let the unbridged edges $\{X_i,X_j\}$, $\{X_k,X_j\}$
be oriented head-to-head by  Principle II. Let 
both edges $\{X_i,X_l\}$, $\{X_j,X_l\}$ or both edges 
 $\{X_k,X_l\}$, 
 $\{X_j,X_l\}$, or all the edges 
 $\{X_i,X_l\}$,
 $\{X_k,X_l\}$, 
 $\{X_j,X_l\}$
 be left unoriented in the 
process. Then orient  $\{X_j,X_l\}$ as  from $X_l$ to $X_j$.
 Apply thereafter Principles II$^c$, IV exhaustively. \\

(If the intrinsic graph is given by Fig.\ref{abbeins} then Principle V 
yields a graph given by Fig.\ref{abbsechs}).\\

\AbbSechs

\begin{th}  \label{iiiDSv} 
Dag-structure and Principle III imply Principle V.
\end{th}

\AbbNeun

\AnfBeweis
The edges 
 $\{X_i,X_l\}$, $\{X_k,X_l\}$
(see Fig.\ref{abbneun})
 are unbridged (because  $\{X_i,X_j\}$, 
$\{X_k,X_j\}$ are unbridged), hence their orientation head-to-head is 
 excluded (as Principle II didn't orient them). Hence either we have 
orientation $( X_l,X_i )$ or  $( X_l,X_k )$\\
 Let us assume the orientation
 $ (X_l,X_i)$ of  $\{X_i,X_l\}$. Then if  $\{X_l,X_j\}$ would be oriented  
$( X_j,X_l )$ then $X_j,X_l,X_i$ would form an oriented  cycle, hence H would 
not be a dag. So this is impossible.\\
 Let us assume the orientation
 $ (X_l,X_k)$ of  $\{X_k,X_l\}$. Then if  $\{X_l,X_j\}$ would be oriented  
$( X_j,X_l )$ then $X_j,X_l,X_k$ would form an oriented  cycle, hence H would 
not be a dag. So this is impossible.\\
Hence $\{X_l,X_j\}$ must  be oriented  
$( X_l,X_j )$ \\
\EndBeweis

We conjecture furthermore that \\
Let $\Gamma$ be the set of directed 
graphs 
that represent DS-belief   distribution Bel according to Principle III. Then 
$\Gamma$ is also the set of directed graphs obtained from P by Principles I 
and II.\\

This conjecture will be proven after showing some intermediate results.\\
We shall introduce first the notion of p-d-separation.

\begin{df}
A {\em p-trail } in a pog is a sequence of links that form a path in the 
underlying  undirected graph. A node $\beta$ is called a head-to-head node 
with 
respect to a p-trail t if there are two consecutive links $\alpha \rightarrow 
\beta$ and $\beta \leftarrow \gamma$ on that t. 
A p-trail is minimal iff no two of its  succeeding links on the p-trail are 
bridged in the graph. 
\end{df}

\begin{df}
A p-descendent of a node n in a pog is any node m such that there exists a 
minimal p-trail from n to m such that every oriented link on the p-trail is 
oriented from n to m and an oriented edge (m,n) does not exist in the graph.
 \end{df}

\begin{df}
A p-trail t connecting nodes $\alpha$ and $\beta$ is said to be {\em active } 
given a set of nodes L, if (1) every head-to-head-node wrt t either is or has 
 a p-descendent in L and (2) every other node on t is outside L. Otherwise t 
is 
said to be {\em blocked } (given L).
\end{df}

\begin{df}
If J,K and L are three disjoint sets of nodes in a pog H, then L is said to 
{\em p-d-separate } J from K, denoted $I(J,K|L)_H$  iff no minimal
p-trail between a 
node in J and a node in K is active given L.
\end{df}

We  claim that 
\begin{th}
Let L be a set of nodes in a pog H, and let $\alpha,\beta \notin L$ be two 
additional nodes in H. Then $\alpha$ and $\beta$ are connected via an active 
p-trail  (given L) iff  $\alpha$ and $\beta$ are connected via a simple (i.e. 
not possessing cycles in the underlying undirected graph) active p-trail 
(given L).
\end{th}


Now let us formulate the central theorem of this paper. 

\begin{th} \label{iuiipiii}
Let D be a dag generated by Principle III. 
Let H be a pog generated by Principles I, II, II$^c$, IV and V.
Then $I(J,K|L)_H$ iff $I(J,K |L)_D$ 
\end{th}

\AbbAcht

\AnfBeweis
To show this, let us consider an  active minimal p-trail. We claim that there 
exists then an active trail. \\
If after final orientation no head-to-head meeting occurs on the p-trail then 
this is also the interesting active trail. Otherwise if there exists a 
head-to-head-meeting on the underlying trail then two cases are possible:
(1) it existed on the original p-trail, (2) it did not exist on the original 
p-trail. The second case is impossible since then it must have been generated 
 by Principle  II (the meeting edges are unbridged). So we have had also a 
head-to-head-meeting on the original p-trail. So let us consider the 
p-descenders of the head-to-head-meeting. No head-to-head-meeting could have 
been generated on the path as the p-trail to the descendent  was minimal. 
p-descendants
of head-to-head meetings
 connected by unoriented links  form a kind of equivalence class in that if 
the edges (A,B), (C,B)  are there and D is a p-descendent of B on a totally 
unoriented path then oriented edges (A,D) and (C,D) are also present. So 
p-descendants are either descendants (OK) or are   such              
     predecessors, that they form together with the nodes of the primary 
p-trail but the discussed head-to-head node
       a minimal p-trail containing that predecessor as a head-to-head node 
and which proves to be an active trail in the dag (see Fig.\ref{abbacht}).\\
Let us consider an active minimal trail. We claim that then there exists a 
minimal p-trail. First of all all the successors are also p-successors. 
Second, a minimal trail is also a minimal p-trail. Now the question  is 
whether 
or not it is also active. As the trail is minimal, no head-to-head meeting 
will vanish on the p-trail. Hence also the successor requirement is met. So 
the proof is complete.
\EndBeweis

This theorem actually corresponds straight forwardly to our
conjecture. The only difference to it is the extensive use of 
Principle II$^c$ which is actually a kind of exploitation of  
Principle II. Furthermore, it is to
 some extent constructive: it states how it 
 is possible to uncover the d-separations applying only Principles I, II,  
II$^c$, IV 
and V for construction of a pog, and without actually instantiating a single 
dag. It is immediately visible, that any  dag compatible with the pog 
expresses 
exactly all the independences  Principle III dag does and hence is a 
Principle III dag.

We can however be still more constructive and formulate the construction
algorithm for generation of all the dags according to Principle III
based only on the results of Principles  I, II,  
II$^c$, IV 
and V and the definition of a dag.

Let us define the legitimate removal of a node from the pog graph:
a node can be removed legitimately from a pog iff all the oriented edges it 
meets are oriented towards it, and all pairs edges meeting at it for which at 
least one is unoriented, are bridged. \\

{\large \bf Pog-to-dag algorithm:}\\
 1. find  a legitimately removable node in the pog, remove it with edges 
meeting it while marking the edges as oriented towards this node. \\
2. Proceed with Step.1 until all the nodes are removed.\\
3. Orient the edges of the original pog so as they were marked in step 1.\\

\AbbSieben

(Compare Fig.\ref{abbsechs}, Fig.\ref{abbsieben}).
We claim that:

\begin{th} 
Let there exist a dag obtainable from  Principle III.
Let G be a pog generated from Principles I, II, II$^c$, IV 
and V.  Then every dag 
obtained from the pog B by the above algorithm 
is a Principle III dag. Every Principle III dag for this population is a dag 
obtainable from G by means of the above algorithm.  
\end{th}
\AnfBeweis
This is easily seen as on the  one hand every dag has a legitimately 
removable node, and on  the other hand the orientations generated by the 
above algorithm do  not  lead  to  any  conflict  with  Principles 
I,II,II$^c$, IV and V, 
if a dag exists.  \\
\EndBeweis

In this way we hope to have also shown the   usefulness
of Theorem \ref{iuiipiii}
definitely, giving a constructive algorithm to generate the dag out of a pog 
which is necessary for belief network applications. \\


This is actually the main result of this paper with respect to
structuring joint 
DS-belief
distributions. It may be stated as follows:\\

\begin{th} 
Let $\Gamma$ be the set of directed 
graphs 
that represent DS-belief   distribution Bel according to Principle III. Then 
$\Gamma$ is also the set of directed graphs obtained from P by Principles I 
and II.
 \end{th}%
\AnfBeweis
Let us look closely at Theorem \ref{iuiipiii}.
From Theorems \ref{iiiDSi} and \ref{iiiDSii} we know that any 
dag D in $\Gamma$ must have been generated also by Principles I and II.
 As Principles II$^c$, 
IV and V follow from Principles I and II and from the property of being a   
dag 
(look at Theorems \ref{iiiDSiic}, \ref{iiiDSiv},    \ref{iiiDSv}), then any 
dag in  $\Gamma$ as generated by Principle III would also be 
generated by Principles I, II, II$^c$, IV and V. Let us take now any of these 
dags in $\Gamma$, say D. Let us assume that from the respective pog H 
generated 
by Principles I, II, II$^c$, IV and V 
(that is in fact from the only such pog H)
a different dag D' may be derived 
beside 
D. From Theorem  \ref{iuiipiii} we have: $I(J,K|L)_H$ iff $I(J,K |L)_D$, but 
also:  $I(J,K|L)_H$ iff $I(J,K |L)_D'$ . Hence also  $I(J,K|L)_D$ iff 
$I(J,K |L)_D'$. But then D' must also have been generated by Principle III as 
both D and D' carry the same independence information.\\
 So we see immediately that any dag in $\Gamma$ must have been generated by 
Principles I and II and all the dags derived via Principles I and II must be 
in $\Gamma$.
\EndBeweis
 
The only open question remains whether a DS-belief distribution can 
be represented using Principle III. \\

\section{When Principle III Fails}

In our theorem on equivalence of d-separation and conditional 
independence 
it is explicitly stated that d-separation is maximal only with respect to 
the whole population of distributions. For a single distribution, there may 
exist independences not covered by the d-separation. Hence for such a 
distribution, Principle III may fail. \\

The non-existence of a dag can in general be visible when:

1. the application of Principle II leads to double oriented edges.\\
2. the oriented subgraph generated by Principle II contains a cycle.\\
3. the application of Principles II$^c$, IV and/or V leads to 
head-to-head-meetings which were forbidden under Principle II.

\AbbZehn

Then no  dag exists  reflecting all the independences for a given population. 
So if we  need  a  dag,  we  have  to  resign  from  some  of  the 
independences. 

\Bem{
 This will affect Principles I and II as 
follows: While applying the Principle I, we will add undirected 
edges between nodes though we find they are independent given 
some set of conditioning variables. While applying Principle II 
we will add head-to-head meetings though conditional 
independence given a set of nodes including the  node shared by 
edges of interest is granted. Let us notice that the application  of 
weakening Principle I will always lead to a dag, but the number of edges in 
the graph will increase. The 
application of weakening  Principle II may not 
 always lead to a dag, but does not increase the number of edges in 
the graph.

The first  two cases need to be resolved by weakening of Principle I. If 
this is 
done, then consequent application off weakening of Principle II will always 
 lead to a dag. Specifically let us turn our attention to the above 
algorithm.   If it fails to find a legitimately removable node, such one has 
to be introduced by means of selection of a node for which all oriented 
coinciding edges are oriented towards it and by forcing the remaining edges 
coinciding with the node to be 
oriented towards it even if the Principle II is 
 violated. \\
}

 The non-existence of a dag may be attributed in the first two cases to 
existence of 
hidden variables, as we can see from examples in Fig.\ref{abbzehn} (double 
orientation of an edge) and in Fig.\ref{abbelf} (a directed cycle). We shall, 
however, not discuss this issue at length here. It can be, however, easily 
checked that introduction of these additional hidden variables as indicated in 

both figures will not give rise to emerging of new independences, not 
present in the population. 

But the third case is hard to resolve. Unless there exist information outside 
the data permitting to assume that the unexpected head-to-head-meeting will 
not introduce unjustified independences in the dag, there may exist the 
necessity to make a complete subgraph out of that part of the graph which 
leads to the unwished head-to-head meeting.

\AbbElf

\section{Summary and Outlook}

In this paper, 
a general framework for recovery of a dag structure of a joint DS-belief
distribution from data has been established, paralleling the work of Spirtes 
et al \cite{Spirtes:90b} on probabilistic networks.
 The proven theorems imply that it is  possible to 
infer causal structure from 
data  if this structure has the form of a directed acyclic graph. Strictly 
speaking: 
The statistical inference allows for deducing a set of such candidate causal 
structures with     indication which fragment of the causal structure is 
shared by 
all the candidates. 

Specifically: In this paper the notion of DS-belief network was introduced 
along with a new notion of conditional independence of variables. 
The applicability of Pearl's notion of d-separation for such a belief
network was demonstrated, especially its relationship to conditional
independence in DS-belief networks. Principles I and II were introduced to 
uncover partial dependency structure of the joint belief distribution.
 An algorithm was given 
allowing 
for derivation of all the dags having identical dependence/independence 
information as the partially oriented graph derived from Principles I and II, 
provided at least one dag exists. If it does not exist, a partial procedure 
transposing the partial dependence structure into one with dag-representation
via introduction of hidden variables 
is also suggested.
 The new notion of p-d-separation paralleling d-separation 
of Geiger, Verma  and Pearl \cite{Geiger:90}, being applicable to partially 
oriented graphs was introduced and has been shown to carry the same 
dependence/independence  information as all the d-separations of all 
compatible dags.

Over the last years a number of alternative methods
to the algorithm of Spirtes 
et al \cite{Spirtes:90b}
 (both general and specialized) 
 for construction of probabilistic belief networks has been proposed  
(compare the method described in  \cite{Cooper:92} and other 
discussed in last 
sections therein). However, they were hardly transferable into
the domain 
of DS-belief functions. For some special case, \cite{Klopotek:93b} offers
solutions.
The method investigated here deserves 
special 
attention because it relates the oriented structure of a directed acyclic 
graph representation to the causal relationship in the described part of 
reality.  Two essential complementary conclusions can be drawn from proving 
theorem \ref{iuiipiii}: 
(i) if one recovers a dag structure for the 
DS-belief   distribution one derives more than just a formal description and 
(ii) for proper construction of a dag causality is essential.
 
Further research on the subject is needed, especially concerning 
approximations binding combinatorial explosion with the number of variables 
considered. 

\newcommand{\ReadingsIn}{G. Shafer, J. Pearl eds: Readings in Uncertain 
Reasoning, (ISBN 1-55860-125-2, 
Morgan Kaufmann Publishers Inc., San Mateo, California, 1990)}

\end{document}